\pdfoutput=1
\documentclass[11pt]{article}

\usepackage[preprint]{coling}

\usepackage{times}
\usepackage{latexsym}

\usepackage[T1]{fontenc}
\usepackage[utf8]{inputenc}

\usepackage{microtype}

\usepackage{inconsolata}
\usepackage{tcolorbox}
\usepackage{graphicx}

\usepackage{float}
\usepackage{latexsym}
\usepackage{amssymb}
\usepackage{amsmath}
\usepackage{amsthm}
\usepackage{booktabs}
\usepackage{enumitem}
\usepackage{subfig}
\usepackage{graphicx}
\usepackage{color}
\usepackage{xcolor} 
\usepackage{hyperref}
\usepackage{appendix} 
\usepackage{tcolorbox}
\usepackage{inconsolata}

\usepackage{multirow}     
\usepackage{colortbl}     
\usepackage{adjustbox}    
\usepackage{booktabs}   
\usepackage{listings}

\definecolor{codegreen}{rgb}{0,0.6,0}
\definecolor{codegray}{rgb}{0.5,0.5,0.5}
\definecolor{codepurple}{rgb}{0.58,0,0.82}
\definecolor{backcolour}{rgb}{0.95,0.95,0.92}

\setlist[itemize]{leftmargin=*}
\setlist[enumerate]{leftmargin=*}
\graphicspath{ {./images/} }

\lstdefinestyle{mystyle}{
  backgroundcolor=\color{backcolour}, commentstyle=\color{codegreen},
  keywordstyle=\color{magenta},
  numberstyle=\tiny\color{codegray},
  stringstyle=\color{blackcolour},
  basicstyle=\ttfamily\footnotesize,
  breakatwhitespace=false,         
  breaklines=true,                 
  captionpos=b,                    
  keepspaces=true,                 
  numbers=left,                    
  numbersep=5pt,                  
  showspaces=false,                
  showstringspaces=false,
  showtabs=false,                  
  tabsize=2
}

\hypersetup{
    hidelinks,
    breaklinks,
    colorlinks,
    linkcolor={blue!70!black},
    citecolor={blue!70!black},
    urlcolor={blue!70!black}
}

\title{Exploring Concept Depth: How Large Language Models Acquire \\Knowledge and Concepts at Different Layers?}

\author{
 \textbf{Mingyu Jin\textsuperscript{\rm 1}},
 \textbf{Qinkai Yu\textsuperscript{\rm 2}},
 \textbf{Jingyuan Huang\textsuperscript{\rm 1}},
 \textbf{Qingcheng Zeng\textsuperscript{\rm 3}},
\\
 \textbf{Zhenting Wang\textsuperscript{\rm 1}},
 \textbf{Wenyue Hua\textsuperscript{\rm 1}},
 \textbf{Haiyan Zhao\textsuperscript{\rm 4}},
 \textbf{Kai Mei\textsuperscript{\rm 1}},
 \textbf{Yanda Meng\textsuperscript{\rm 2}},
\\
 \textbf{Kaize Ding\textsuperscript{\rm 3}},
 \textbf{Fan Yang\textsuperscript{\rm 5}},
 \textbf{Mengnan Du\textsuperscript{\rm 4}},
 \textbf{Yongfeng Zhang\textsuperscript{\rm 1}}
\\
 \textsuperscript{1}Rutgers University,
 \textsuperscript{2}University of Exeter,
 \textsuperscript{3}Northwestern University,\\
 \textsuperscript{4}New Jersey Institute of Technology,
 \textsuperscript{5}Wake Forest University,
\\
 \small{
 \texttt{\{mingyu.jin, chy.huang, yongfeng.zhang\}\href{mailto:@rutgers.edu}{@rutgers.edu}}
 },
 \small{
 \texttt{{qy269}\href{mailto:@exeter.ac.uk}{@exeter.ac.uk}}
 },
 \small{
 \texttt{{mengnan.du}\href{mailto:@njit.edu}{@njit.edu}}
 },\\
}

\begin{document}
\maketitle
\begin{abstract}
Large language models (LLMs) have shown remarkable performances across a wide range of tasks. However, the mechanisms by which these models encode tasks of varying complexities remain poorly understood. In this paper, we explore the hypothesis that LLMs process concepts of varying complexities in different layers, introducing the idea of ``Concept Depth'' to suggest that more complex concepts are typically acquired in deeper layers. Specifically, we categorize concepts based on their level of abstraction, defining them in the order of increasing complexity within factual, emotional, and inferential tasks. We conduct extensive probing experiments using layer-wise representations across various LLM families (Gemma, LLaMA, Qwen) on various datasets spanning the three domains of tasks. Our findings reveal that models could efficiently conduct probing for simpler tasks in shallow layers, and more complex tasks typically necessitate deeper layers for accurate understanding.
Additionally, we examine how external factors, such as adding noise to the input and quantizing the model weights, might affect layer-wise representations. Our findings suggest that these factors can impede the development of a conceptual understanding of LLMs until deeper layers are explored. We hope that our proposed concept and experimental insights will enhance the understanding of the mechanisms underlying LLMs. Our codes are available at \url{https://github.com/Luckfort/CD}.
\end{abstract}

\renewcommand{\thefootnote}{\fnsymbol{footnote}}
\footnotetext{**Mingyu, Qinkai, Jingyuan, and Qingcheng are the main contributors. The order of these authors is determined by flipping a coin.**\\
\url{https://luckfort.github.io/explore\_CD/} is the project website.}
\section{Introduction}
\begin{figure*}[h]
\centering
\includegraphics[width=0.95\textwidth]{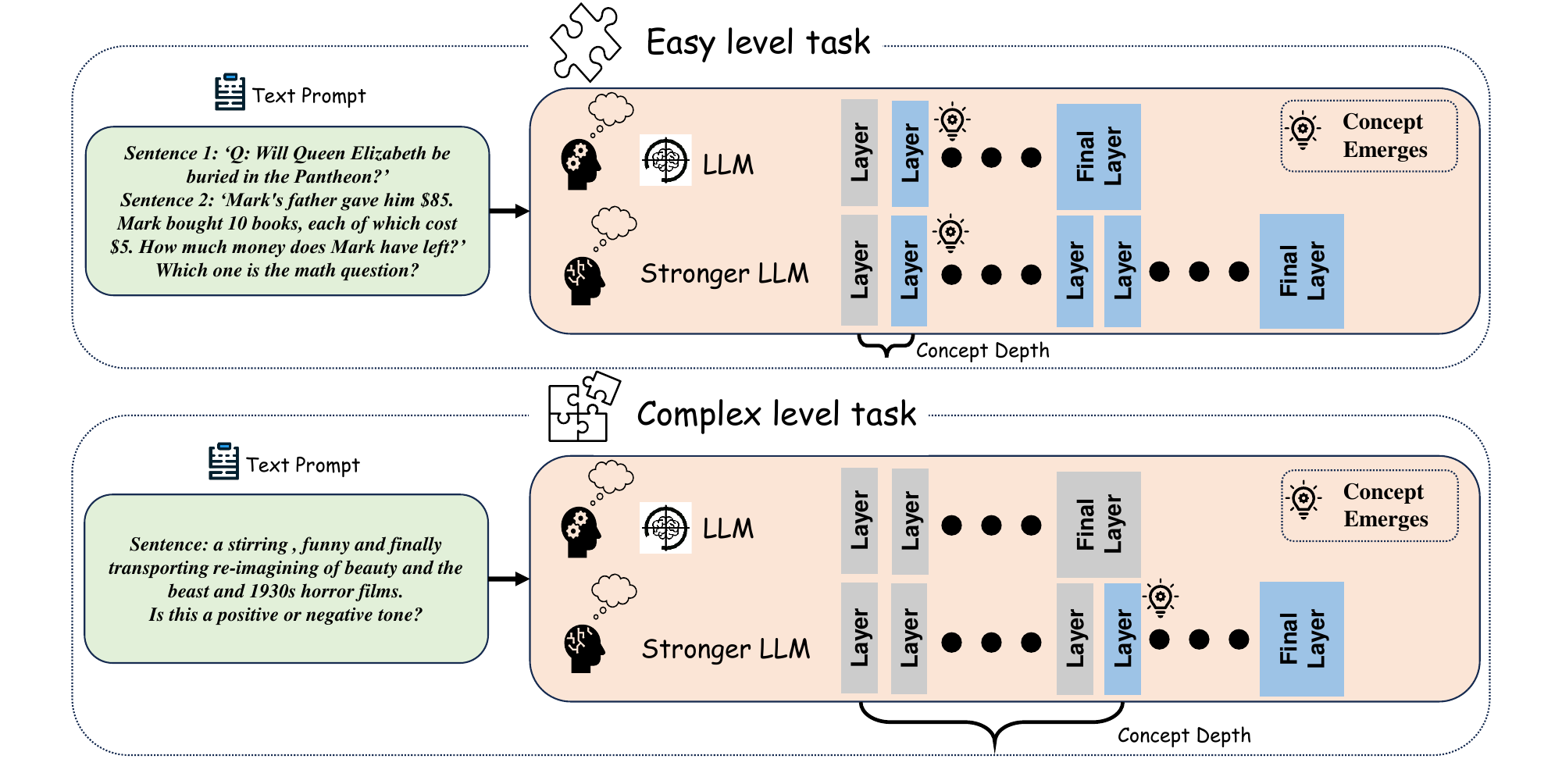}
\caption{The LLMs are trying to understand easy tasks and complex tasks. The more complex the task, the deeper layers an LLM needs to understand. The stronger the LLM is, the more complex the task level it can learn.}
\vspace{-0.3cm}
\label{fig:task}
\end{figure*}

LLMs such as GPT-4~\cite{achiam2023gpt} and LLaMA~\cite{touvron2023llama} have impressive generation and reasoning capabilities~\cite{chang2023survey, su2024living, su2024timo}. It is widely accepted that these models embed substantial knowledge in their parameters, with performance improving as the number of parameters increases~\cite{ju2024large,jin2024disentangling,zhou2024mathattack}, also known as emergent abilities~\cite{wei2022emergent}. For instance, GPT-3~\cite{brown2020language} shows a large increase in performance after scaling up to 13B parameters, and a similar phenomenon was also observed for LaMDA~\cite{thoppilan2022lamda} after exceeding 68B parameters~\cite{wei2022chain}. However, it is not well understood how LLMs accurately grasp the concept of knowledge. In this paper, we investigate the following research question: \emph{Can shallow layers in LLMs capture meaningful features of simple knowledge, while complex concepts need deeper layers to capture their meaningful features?} like \autoref{fig:task}. We hope to explore the connection between the depth of language models' neural networks and their conceptual understanding ability by studying this question.

\begin{figure*}[!t]
	\centering 
\includegraphics[width=0.95\linewidth]{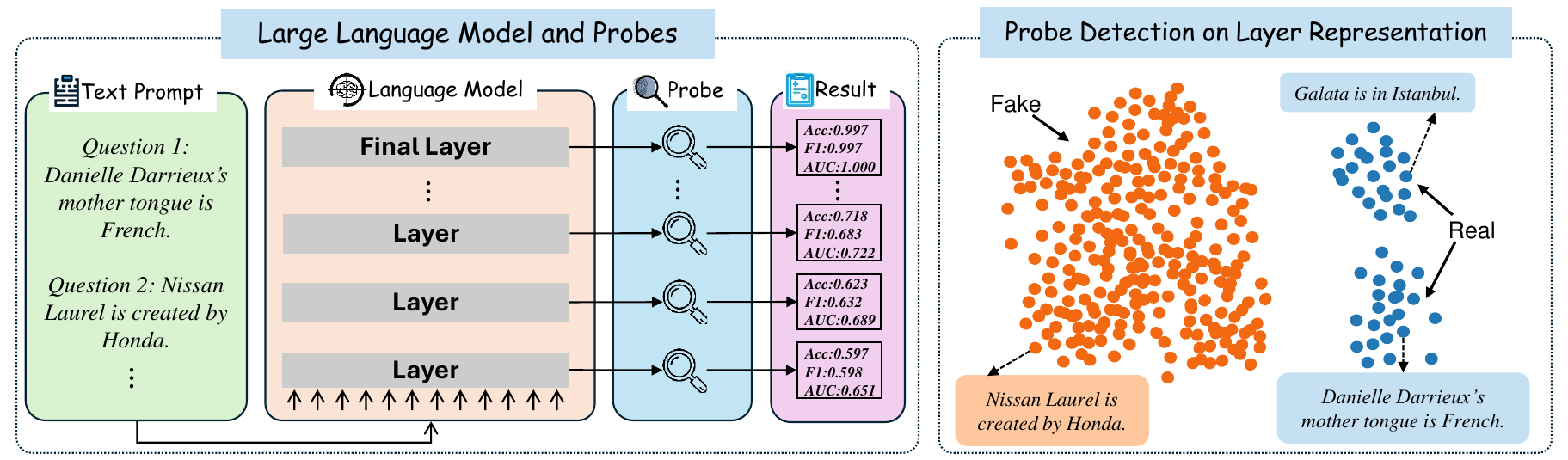}
\caption{The left figure provides an overview of our analysis process. LLMs respond to text prompts, and the probing process assesses the optimal performance achievable by the current LLM layer. The right figure illustrates a demonstration of layer-wise representations by probe detection. In this demonstration, orange points represent fake samples, while blue points represent real samples. In this case, the probe tries to classify between two categories.
}
\vspace{-0.3cm}
\label{fig:overall}
\label{figure1}
\end{figure*}

Recent studies on understanding the reasoning abilities of LLMs focus on two main strategies: probing representations and model pruning. Probing involves using linear classifier probes to analyze the performance of hidden layer representations; for instance,~\citet{duan2024llms} examines changes in LLMs' internal representations during hallucinations, while~\citet{ju2024large} investigates the performance of different layers in the LLaMA series using synthetic counterfactual datasets. On the other hand, model pruning removes redundant parameters based on their importance in seeing if performance is significantly affected. This method, although effective, can be complex and time-consuming. For example,~\citet{zhang2023pruning} uses gradient information to decide the pruning components, and~\citet{gromov2024unreasonable} even requires QLoRA fine-tuning~\cite{dettmers2024qlora} to do the pruning.
Given these complexities, Our work primarily analyzes the representations obtained through probing techniques. Building upon previous work, we aim to gain a more comprehensive understanding of the layer representations within LLMs.

Our general framework is shown in~\autoref{fig:overall}. We trained independent linear probes for each layer of LLMs to predict the binary label, thereby determining the optimal performance achievable with the representations of each respective layer. Drawing from our empirical findings, we propose the notion of ``Concept Depth'' as a novel metric to evaluate the capability of different models in comprehending varying levels of knowledge across their layers. This is the first time such a concept has been introduced in the relevant literature.

Our empirical results ranging from 3 popular LLMs families (Gemma~\cite{team2023gemini}, LLaMA~\cite{touvron2023llama}, and Qwen~\cite{bai2023qwen}) and 9 datasets reveal that ``Concept Depth'' is widely applicable in existing mainstream LLMs. Besides, we conducted comprehensive robustness analyses, introducing random strings as the noise or quantization, to further understand how the reasoning of LLMs is sensitive to noise. 
To conclude, our main contributions could be summarized as follows:

\iffalse
\begin{figure}[!t]
\begin{center}
    \includegraphics[width=0.45\textwidth]{img/Picture1.pdf}
    \caption{The overview of the process of our analysis. LLMs respond to text prompts. The probing process evaluates the optimal performance that the current layer of LLMs can achieve.}
    \vspace{0.2cm}

    \label{fig:General1}
\end{center}
\end{figure}
%%
\fi
{\fontsize{10}{7}\selectfont\textbullet}
\noindent \emph{Concept Depth.} We introduce the idea of ``Concept Depth'' to measure different layers' abilities to learn different levels of concepts. We first anchored the difficulty of the dataset using LLaMA-3-8b-Instruct~\cite{dubey2024llama} and then tested ``Concept Depth'' with other models. Our results show that simpler concepts are often learned at shallower levels, while complex concepts require deeper levels to understand like Demo \ref{fig:task}. This phenomenon has been observed across LLMs of different model families and different sizes.

{\fontsize{10}{7}\selectfont\textbullet}
\noindent \emph{Experiments on understanding capabilities of LLMs.} We experimented with multidimensional datasets (fact, emotion, and reasoning) to analyze variations in the conceptual depth of LLMs. We observed these differences across various datasets, model parameter counts, and model families (Gemma~\cite{team2023gemini}, LLaMA~\cite{touvron2023llama}, and Qwen~\cite{bai2023qwen}), providing a concise understanding of their impact on LLM performance and comprehension. 

{\fontsize{10}{7}\selectfont\textbullet}
\noindent \emph{Robustness from Concept Depth perspective}.
 We provide a new perspective on LLMs robustness. We conduct ablation experiments on model weight quantization and add random noise to the input that may affect the accuracy of LLMs inference. Details can be found in Appendix \ref{sec:ablastudy}. The results show that after adding the noise or conducting the quantization on the weights, the LLMs end up learning the concepts at slower paces and deeper layers.

\section{Related Work}

\subsection{Concepts Representation in DNNs}
Identifying similarities across various examples to form concepts, plays a crucial role in human decision-making~\cite{armstrong1983some}. Many studies have explained DNNs' (Deep Neural Networks) decision-making based on a conceptual perspective, describing the global behavior of DNNs in a human-understandable way~\cite{su2022improving, raz2023methods, ren2023defining, deng2021discovering,wen2024gaprotonet,wen2024language,wen2024impact}. For example,~\cite{yeh2019concept} demonstrated that DNNs exhibit conceptual representations through the activation patterns observed in their hidden or output layers. Further,~\citet{raz2023methods} indicates that DNNs learned not only conceptual representations of predicted categories but also indirect concepts that contribute to the prediction. A notable study reveals the existence of a representation bottleneck, highlighting a cognitive disparity between DNNs and humans. This phenomenon is characterized by DNNs' tendency to grasp representations of concepts that are either too simple or overly complex, while they often struggle with acquiring representations of concepts of moderate complexity~\cite{deng2021discovering}. Motivated by previous work, our paper aims to cover concepts of different complexities to understand concepts within LLMs further.

\subsection{Knowledge and Concepts in LLMs}
The impressive performance of the LLMs in various downstream tasks (\textit{e.g. LLMs can predict factual statements about the world based on prompts}~\cite{meng2022locating}) has led to a great discussion about whether these capabilities are `stochastic parrots' or LLMs understands these concepts.
Pioneeringly,~\citet{gurnee2023language} showed that LLMs internally store concepts like latitude, longitude, and time. Similarly, another work showed that the internal states of LLMs can detect the truth of a statement~\cite{azaria2023internal, su2024conflictbank}.~\citet{geva2023dissecting} also came to similar conclusions by artificially blocking or ``knocking out" specific parts of the LLMs to observe their effects on the inference process. These related studies show the existence of structures for understanding concepts within LLMs, motivating us to explore how concepts at various complexities are encoded within various depths of LLMs.
 
\subsection{Explorations of Interpretability in LLMs}

Many related studies have deconstructed the inner layers of LLMs from various perspectives to understand the mechanisms inside such models~\cite{zhao2024opening,jin2024impact}.~\citet{fan2024not} computes stopping signals by evaluating key features to early stop the LLM inference and get the internal performance of the LLMs, concluding that not all layers are necessary. Through pruning the LLMs,~\citet{gromov2024unreasonable} found that the parameters of some layers were not utilized correctly.~\citet{men2024shortgpt} also shows a high level of redundancy in the LLMs' architecture. Probes trained with logistic regression are a well-established method~\cite{alain2016understanding} that has been applied in classifying the truthfulness of LLMs and has been validated in many studies~\cite{marks2023geometry, azaria2023internal, li2024inference}. The latest work detects different layers in the Llama series responding to facts or counterfactuals by probing techniques~\cite{ju2024large}. Inspired by these works, we propose Concept Depth to summarize these phenomena. Our work focuses on the Concept Depth that appears in the LLMs, analyzing it experimentally by training linear classifier probes, which makes our work different from others.

\section{Analyzing Method}

In this paper, we design a probing framework to understand how concepts at various levels are encoded within LLMs and investigate whether the internal representations are robust to concepts. For instance, \autoref{fig:overall} demonstrates the representation project of the Counterfact dataset.

\subsection{Linear Classifier Probing}
Probe technology~\cite{alain2016understanding} is a method for analyzing and evaluating the internal representations of a neural network by applying a specific probe task, typically a classification or regression task, to a particular layer of the model. This technique measures the layer's ability to represent information for the given task, thereby revealing the features and information captured by different layers of the model. Our approach involves extracting the representations from each layer of the large model, training a binary classifier on these representations, and validating its accuracy.

For one specific task $w$ that contains $n$ questions, the hidden feature set in LLMs is $x \in \mathbb{R}^{n \times d_{model}}$, where $n$ denotes number of samples, and $x^{(i)} \in \mathbb{R}^{1\times d_{model}}$ represent the representation at a certain layer, where $d_{model}$ donate the dimension for the hidden layer representation. Binary label $y^{(i)}$ is set as 0 or 1.  The objective function of such a binary Logistic regression classifier probes with L2 regularization can be written as:
\begin{equation}
\scriptsize
%J(\theta) = -\frac{1}{m} \sum_{i=1}^{m}[y_{(i)}\log g_{\theta}(H^{(i)})+\sum_{i}^{m}\log(1-g_{\theta}(H^{(i)}))]
   J(\theta) = -\frac{1}{n} \sum_{i=1}^{n} Cost(\sigma(x^{(i)}),y^{(i)})+\frac{\lambda}{2n}\sum_{j=1}^{m}\theta_{j}^2
\end{equation}
\begin{equation}
\scriptsize
\begin{array}{cc}
     &Cost(\sigma(x^{(i)}),y^{(i)}) = y^{(i)} \log \left( \sigma(\theta^T x^{(i)}) \right) \\
     &      + (1 - y^{(i)}) \log \left(1 - \sigma(\theta^T x^{(i)}) \right) 
\end{array}
    % Cost(\sigma(x^{(i)}),y^{(i)}) =
    % y^{(i)} \log \left( \sigma(\theta^T x^{(i)}) \right) + (1 - y^{(i)}) \log \left(1 - \sigma(\theta^T x^{(i)}) \right) 
\end{equation}
where $\theta$ is the parameter in this Logistic regression model, $\lambda$ is the regularization parameter. The linear model predicts LLM's response to the test set, compared with the true label. This yields a quantification of how well LLMs understand the current depth. If the binary model gets good accuracy at a certain layer, that means the LLM can distinguish true or false in this layer.

\section{Experimental Setting}
Our experiments used nine datasets containing three aspects (emotion understanding, reasoning, and fact-checking). We categorized these nine datasets from easy to complex levels according to the performance of LLaMA3-8B-Instruct~\cite{dubey2024llama}, GPT-4o-mini~\cite{gpt4o}, and QWen2-7B-Instruct~\cite{yang2024qwen2} on each dataset (see Section \ref{Anchoring}) to anchor the difficulty of the datasets. Specifically, the datasets in which the linear probes can obtain high classification accuracy at the initial or middle depth of the LLMs are categorized as easy levels. Other datasets where linear probes can only accurately classify at a deeper layer of the model or even fail to classify accurately are categorized as complex levels. The average accuracy of these datasets on the three models was consistent with the probe results and had a significant correlation. In Section \ref{sec:Models}, we introduce the LLMs used for experiments. The nine datasets are described in Section \ref{sec:dataset}. 

\begin{figure*}[!ht]
\begin{center}
    \includegraphics[width=0.90\textwidth]{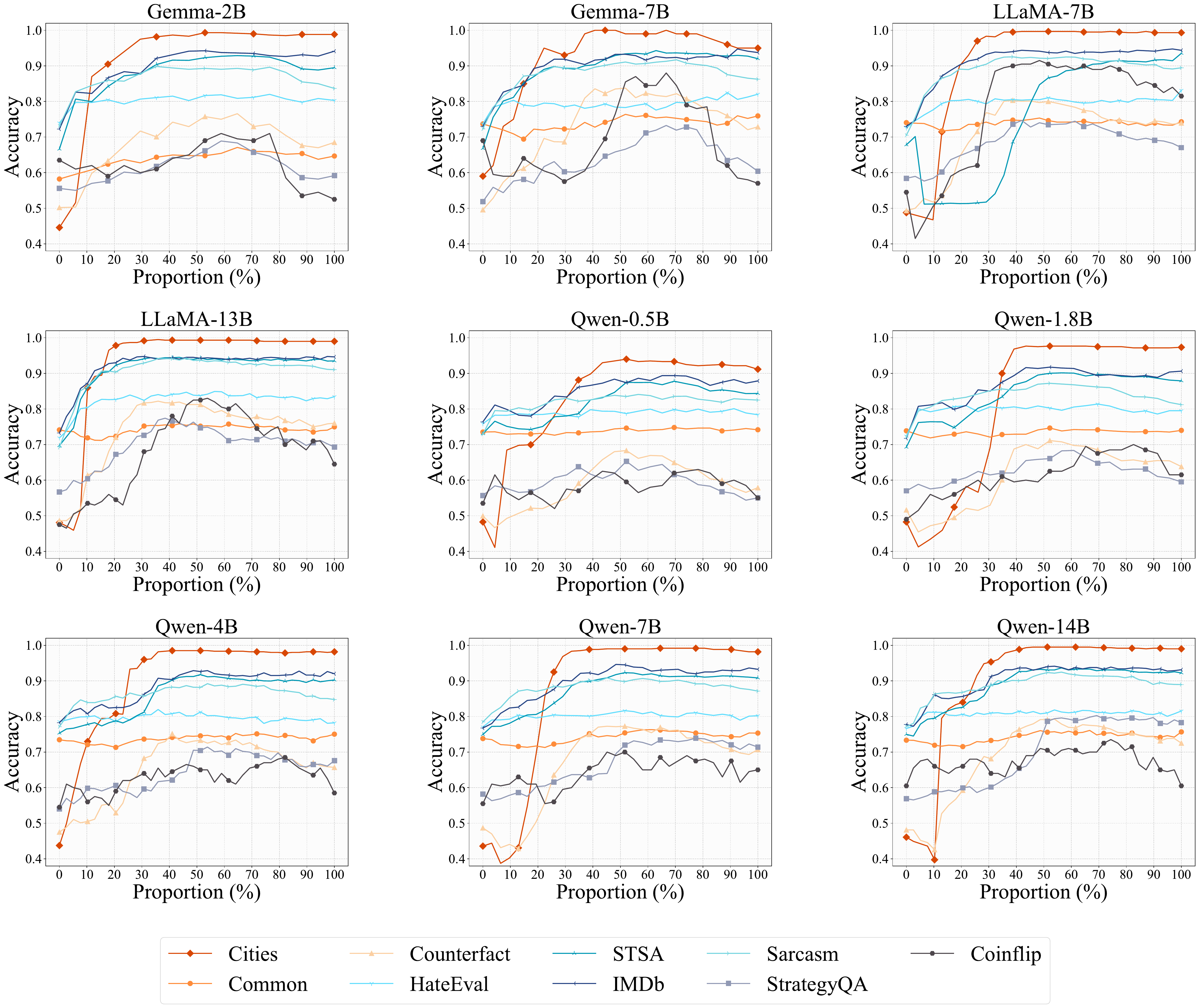}
    \caption{Analysis diagrams of Section \ref{sec:comparedataset}. Linear probing accuracy of three LLM families (Gemma, LLaMA, Qwen) on nine datasets.}
    \label{abla:diffdataset}
\end{center}
\vspace{-0.4cm}  
\end{figure*}

\subsection{Models}
\label{sec:Models}
In this paper, we employ three open-source model families: Gemma (2B, 7B)~\cite{gemmateam2024gemma}, LLaMA-2 (7B, 13B)~\cite{touvron2023llama}, and Qwen (0.5B, 1.8B, 4B, 7B, and 14B)~\cite{bai2023qwen} to support our analysis.  \autoref{tab:numlayer} shows the number of layers in each model. We choose a linear classifier probe for the experiments during the probing analysis. The ratio of the training set to the test set is 8:2, following the usual approach of LLMs probing classifier~\cite{duan2024llms, pal2023future}. We extract feature representations from the final layer in the transformer at each layer of LLMs(\textit{e.g.} \textit{14-th 'post\_attention\_layernorm' in Llama2-7B (32 Layers in total)}) as input to the probing classifier. The other series of models follow a similar processing pattern.

\begin{table}[h]
\scriptsize
%\vspace{0.1cm}
\renewcommand\arraystretch{1.3}

\centering
\begin{tabular}{l l l l l l }
\hline
Model & Layer & Model & Layer & Model & Layer\\
\hline
Gemma-2B &18& Qwen-4B &40& LLaMA-7B &32\\
Qwen-0.5B &24& Gemma-7B &28& Qwen-14B &40\\
Qwen-1.8B &24& Qwen-7B &32& LLaMA-13B &40 \\
\hline
\end{tabular}
\vspace{-0.3cm}
\caption{Number of layers in each LLM.}
\label{tab:numlayer}
\end{table}

\subsection{Datasets}
\label{sec:dataset}
% We categorized our datasets into low-level task datasets and high-level task datasets. 
\autoref{tab:datasetcate} presents the nine datasets we use, on \textbf{Fact} (Cities~\cite{marks2023geometry}, CommonClaim~\cite{casper2023explore}, Counterfact~\cite{meng2022locating}), \textbf{Emotion} (STSA~\cite{kim-2014-convolutional}, IMDb~\cite{maas-EtAl:2011:ACL-HLT2011}, Sarcasm~\cite{misra2023Sarcasm}, HateEval~\cite{manolescu-etal-2019-tueval}), and \textbf{Reasoning} (StrategyQA~\cite{10.1162/tacl_a_00370}, Coinflip~\cite{wei2022chain}) for our experiments. A detailed description of the dataset can be found in the Appendix \ref{adata}.

\subsubsection{Anchoring Difficulties of Each Dataset}
\label{Anchoring}
To ascertain the learning difficulty of each dataset, we have utilized the LLaMA3-8B-Instruct~\cite{dubey2024llama}, GPT-4o-mini~\cite{gpt4o}, and QWen2-7B-Instruct model~\cite{yang2024qwen2}. Our approach involves testing each sample in the datasets as a binary classification problem via a prompting way. The model generates a response for each sample, from which we infer a judgment, categorizing it as either "Yes" or "No". By comparing these judgments with the actual labels, we compute the accuracy for each dataset.

\autoref{tab:anchor} presents the results of this analysis. The dataset with the highest accuracy is deemed the easiest dataset to classify. Conversely, the dataset with the lowest accuracy is considered the most difficult to classify. This method quantitatively measures the learning difficulty associated with each dataset.

\begin{table}[tb]
\scriptsize
%\vspace{-0.5cm} % 调整表格与上一段文字的距离
\renewcommand\arraystretch{1.3}

\centering
\resizebox{0.23\textwidth}{!}{
\begin{tabular}{l l c l}
\toprule
 & Dataset & Accuracy\\
\midrule
 &Coinflip & \cellcolor{cyan!31}0.5920&  \\
 &Common & \cellcolor{cyan!35}0.6487&  \\
 &Sarcasm & \cellcolor{cyan!36}0.6597&  \\
 &StrategyQA & \cellcolor{cyan!46}0.6969&  \\
 &Counterfact & \cellcolor{cyan!50}0.7126&  \\
 &HateEval & \cellcolor{cyan!56}0.7640&  \\
 &STSA & \cellcolor{cyan!69}0.9116&  \\
 &Cities & \cellcolor{cyan!70}0.9204&  \\
 &IMDb & \cellcolor{cyan!74}0.9380&  \\

\bottomrule
\end{tabular}}
\caption{Average accuracy on nine datasets based on LLaMA3-8b-Instruct, GPT-4o-mini and QWen2-7B-Instruct. Accuracy based on each model is shown in \autoref{tab:detailanchor}.}
\label{tab:anchor}
%\vspace{-1cm} % 调整表格与下一段文字的距离
\end{table}

\subsection{Metrics for Accuracy Variation}
% We introduce two metrics to capture variations in accuracy: the jumping point and the converging point.\\
\textbf{Definition 1} \textit{(Variation Rate) Given an LLM probing classifier $M=\{q, y, z, d\}$ ($q$, $y$, $z$, and $d$ are the input question, ground truth binary label, predicted label and total amount of layers, respectively), it has the accuracy $\alpha_i$ at i-th layer: $$\alpha_i=\frac{1}{|z|}*\sum_{k=1}^{|z|}{[y_k=z_k]}, i \in \{0,1,2,...,d-1\}$$ We denote the variation rate $\beta_i$ where $$\beta_i=\alpha_i / \alpha_{i-1}, i \in \{1,2,...,d-1\}$$}

We introduce two metrics to capture variations in accuracy: the \textbf{jumping point} and the \textbf{converging point} and define them by the given definition of variation rate.

\textbf{Definition 2} \textit{(Jumping point) We denote the jumping point $$J(M, D)=\min\{\frac{i}{d}\}$$ $$\ s.t.\ \beta_{i}>=1.1, i \in \{1,2,...,d-1\}$$ where M and $D=(q, y)$ are the LLM classifier and the dataset.} 

When a noticeable boost in accuracy is observed, the jumping point signals the model's recognition of a dataset's critical patterns.

\textbf{Definition 3} \textit{(Converging Point) We denote the converging point $$C(M, D)=\max\{\frac{i}{d}\}$$ $$s.t. |\beta_{i} - 1|<0.03, i \in \{1,2,...,d-1\}$$ where $M$ and $D=(q, y)$ are the LLM classifier and the dataset.}

% needs revising
As the accuracy plateaus or starts declining, the converging point indicates the model's learning saturation or peak learning capacity from the dataset. Analyzing these metrics offers deeper insight into the model's learning dynamics and adaptability to various data types.

\begin{figure*}
\begin{minipage}[b]{1\linewidth}
    \centering
    \subfloat[][The converging point of each dataset on Gemma, LLaMA, and Qwen represented by the percent depth proportion.]
    {\label{p2:gl}\includegraphics[width=1\textwidth]{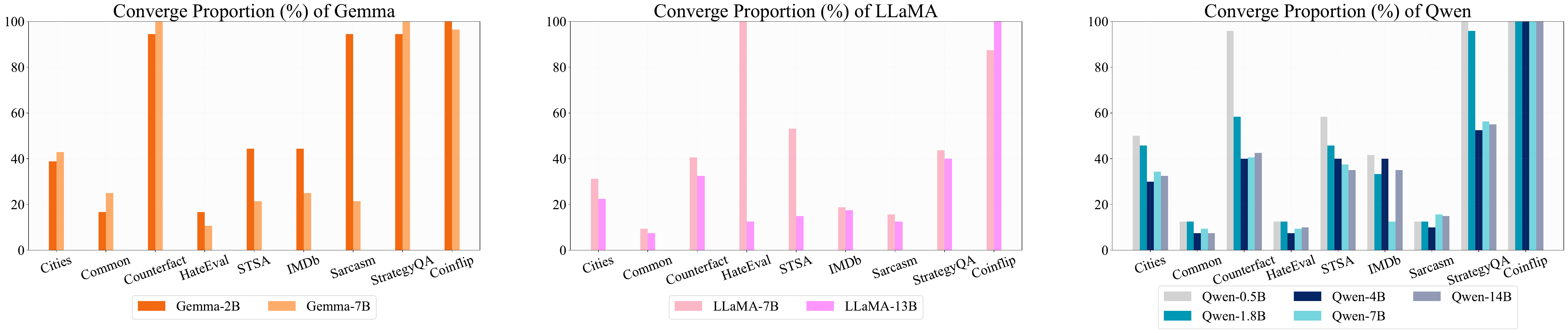}\vspace{0.2cm}}

    \subfloat[][The peak accuracy of each dataset on Gemma, LLaMA, and Qwen represented by the percent depth proportion.]
    {\label{p2:gl1}\includegraphics[width=1\textwidth]{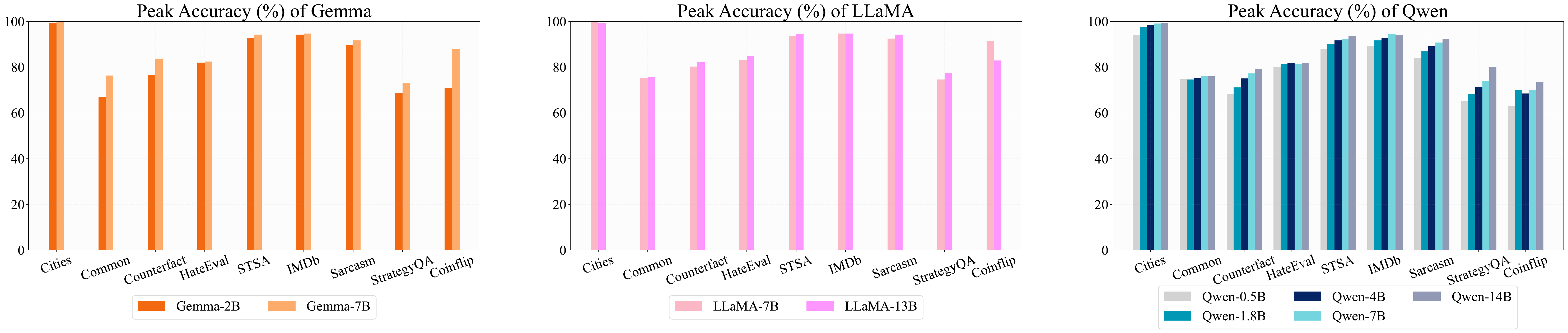}\vspace{0.2cm}}
\end{minipage} %\par
\caption{Analysis diagrams of Section \ref{sec:comparediffB}. The converge proportion and peak accuracy of each model over the nine datasets. (a) shows the converged proportion over the datasets. (b) shows the peak accuracy over the datasets.}
\label{fig:52}
\vspace{-0.1cm}
\end{figure*}

\begin{figure*}[!t]
\begin{center}
    \includegraphics[width=0.87\textwidth]{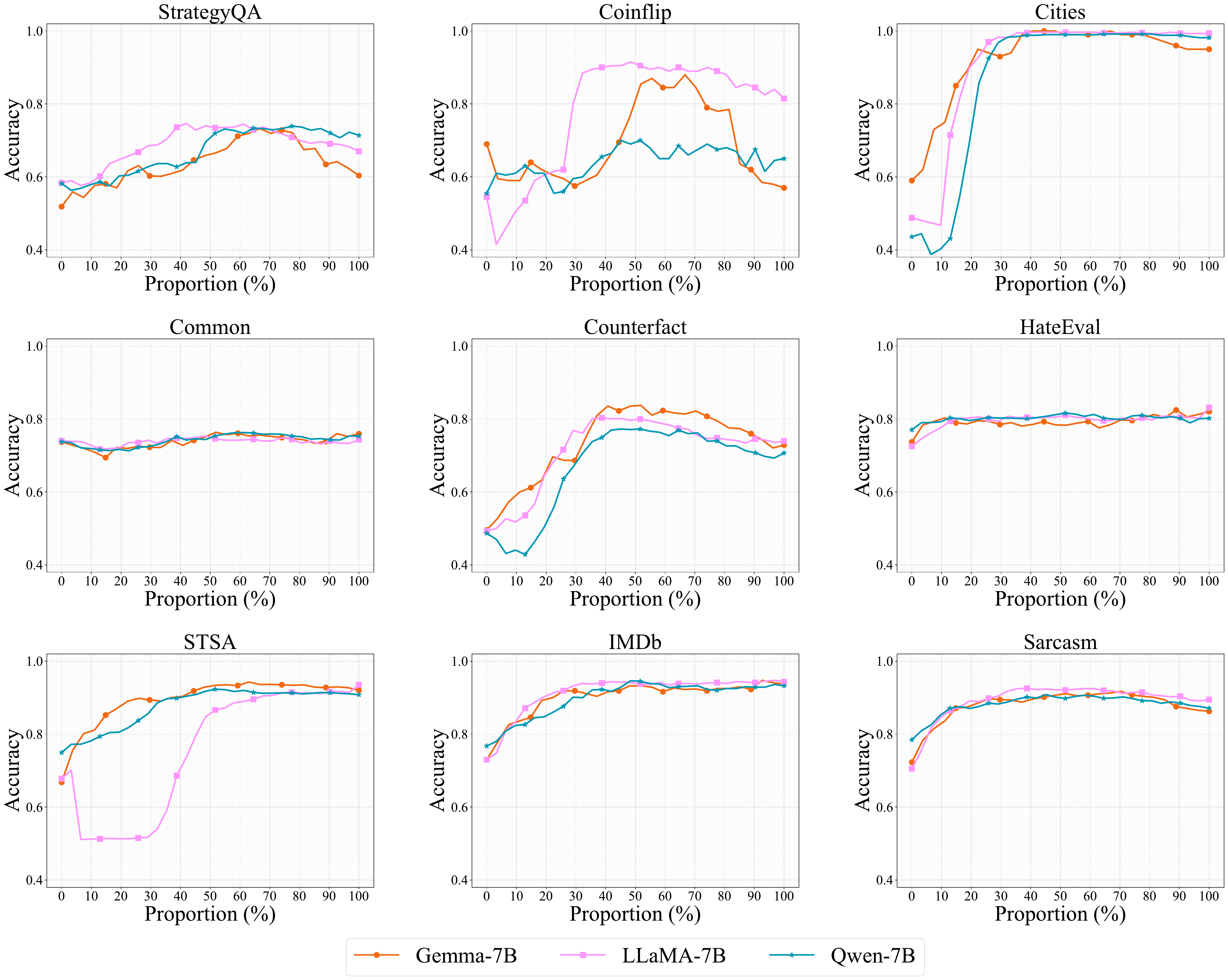}
    \caption{Analysis diagrams of Section \ref{sec:comparediffFMLY}. Linear probing accuracy of Gemma-7B, LLaMA-7B, Qwen-7B on nine datasets.}
    \label{7b:general}
\end{center}
\vspace{-0.5cm}  
\end{figure*}

\section{Experimental Analysis}
We conduct experiments to answer the following research questions about the Concept Depth:

 \textbf{RQ1}: Do different LLMs' Concept Depths behave consistently in the same dataset? (Section \ref{sec:comparedataset})

 \textbf{RQ2}: Do different size LLMs in the same family (\textit{e.g., the LLaMA family}) have consistent Concept Depth? (Section \ref{sec:comparediffB}) 

 \textbf{RQ3}: Do LLMs' Concept Depth of the same size behave consistently? (Section \ref{sec:comparediffFMLY})
\subsection{Comparison Among the Datasets}
\label{sec:comparedataset}
We delve into an evaluative performance comparison across a range of datasets, utilizing \autoref{abla:diffdataset} to detail the layer-wise accuracy of all nine LLMs over nine distinct datasets. \autoref{tab:stat} shows the detailed numerical results for \autoref{abla:diffdataset}, as well as the F1-score and AUC. A performance threshold of 0.7 accuracy is applied to assess the models' effective comprehension of concepts. This examination leads to two general observations. Firstly, regarding different concepts, LLMs exhibit varying accuracy trends across their layers. For example, \textbf{Cities} approaches perfect accuracy fast; in contrast, datasets requiring high-level reasoning such as \textbf{StrategyQA} will not reliably converge to accuracy above 0.7, indicating that they have different ``Concept Depth''. Within individual concepts, however, different LLMs tend to display consistent accuracy patterns across these layers. Secondly, in tasks that require varying levels of conceptual understanding, the LLMs demonstrate their understanding across different layers, indicating a layered approach to processing complex concepts.

 Significant variations in trends are observed across the models among the three factual concept datasets. \textbf{Cities} exhibits a sharp increase in comprehension at lower layers, stabilizing in higher layers, indicating a strong grasp of the concept. \textbf{CommonClaim} has become stable in early layers. 
Besides, the accuracy improvement of the nine LLMs trained on \textbf{Counterfact} was relatively difficult to achieve, utilizing deeper layers, and the accuracy was lower than that of many other datasets. Therefore, we can conclude that \textbf{Counterfact} is more complex.

In datasets centered on emotional concept comprehension (\textbf{STSA}, \textbf{IMDb}, \textbf{Sarcasm}, and \textbf{HateEval}), despite varying levels of understanding, all models demonstrate a rise in accuracy at the initial layers, with convergence occurring in the intermediate layers. Although \textbf{HateEval} essentially reaches stable at the initial layers, its accuracy reaches up to 0.8, suggesting that LLMs primarily aggregate representations from lower layers to grasp emotional concepts. Meanwhile, \textbf{StrategyQA} and \textbf{Coinflip}, which demand specific reasoning skills, tend to display a bell-shaped accuracy trajectory in all models, with peak accuracy observed in the middle layers. Such patterns underscore the intricate complexity associated with reasoning tasks.
\begin{tcolorbox}[colback=gray!10!white,colframe=gray!70!black,boxrule=0.3pt,title=Remark 1]
We categorize the performances into three types. 1) For \textbf{Cities}, \textbf{STSA}, \textbf{IMDb},  and \textbf{Sarcasm}, the LLMs suddenly understand the tasks at intermediate layers. 2) For \textbf{CommonClaim} and \textbf{HateEval}, the LLMs have already understood the tasks in shallower layers. 3) For \textbf{Counterfact}, \textbf{StrategyQA}, and \textbf{Coinflip}, The tasks are more difficult to understand compared with others. Therefore, we consider the tasks in type 1 and 2 easy tasks, and those in type 3 are complex. 
\end{tcolorbox}
\subsection{Comparison Among the Number of Parameters}
\label{sec:comparediffB}
This section offers a comparative analysis of LLMs within their respective families, examining both accuracy levels and converging points across the models. \autoref{fig:52} reveals two recurring patterns within these families: for tasks with accuracy improves dramatically by model learning, larger models tend to show converging points at earlier layers, suggesting they achieve their own peak comprehensions of concepts at lower layers; for tasks with accuracy changes little, all LLMs show the converging points at early layers. 

Two notable exceptions to this trend appear in the Qwen family over the \textbf{Coinflip} and \textbf{IMDb} datasets. For \textbf{Coinflip}, larger models exhibit delayed convergence. This deviation underscores the complexity of the reasoning required, illustrating how this task challenges even the larger models to extend their depth of understanding further. For \textbf{IMDb}, converging points fluctuate with the increasing size of the model because the number of layers is different among different sizes of LLMs, which amplifies the differences. These exceptions are also found in the Gemma family.

Furthermore, in \autoref{fig:52}, we explore the peak accuracy levels across all layers for LLMs of differing sizes. The overarching trend indicates that larger models consistently achieve superior peak performance. This observation not only supports that scaling up models enhances their effectiveness but also suggests that larger models develop more robust internal representations, validating the benefits of training models with greater capacity.

\begin{tcolorbox}[colback=gray!10!white,colframe=gray!70!black,boxrule=0.3pt,title=Remark 2]
We have two observations by comparing different sizes of models from the same LLM family. 1) As the number of parameters increases, peak accuracy gradually increases, and the converging point gradually advances. 2) Larger models grasp the concepts earlier and better.
%Increasing model parameters leads to gradual improvements in peak accuracy and earlier convergence. Larger models grasp concepts more effectively.
\end{tcolorbox}
\subsection{Comparison Among the LLM Families}
\label{sec:comparediffFMLY}
\begin{figure}[!t]
    \hspace{-0.1cm}
    \includegraphics[width=1.1\textwidth]{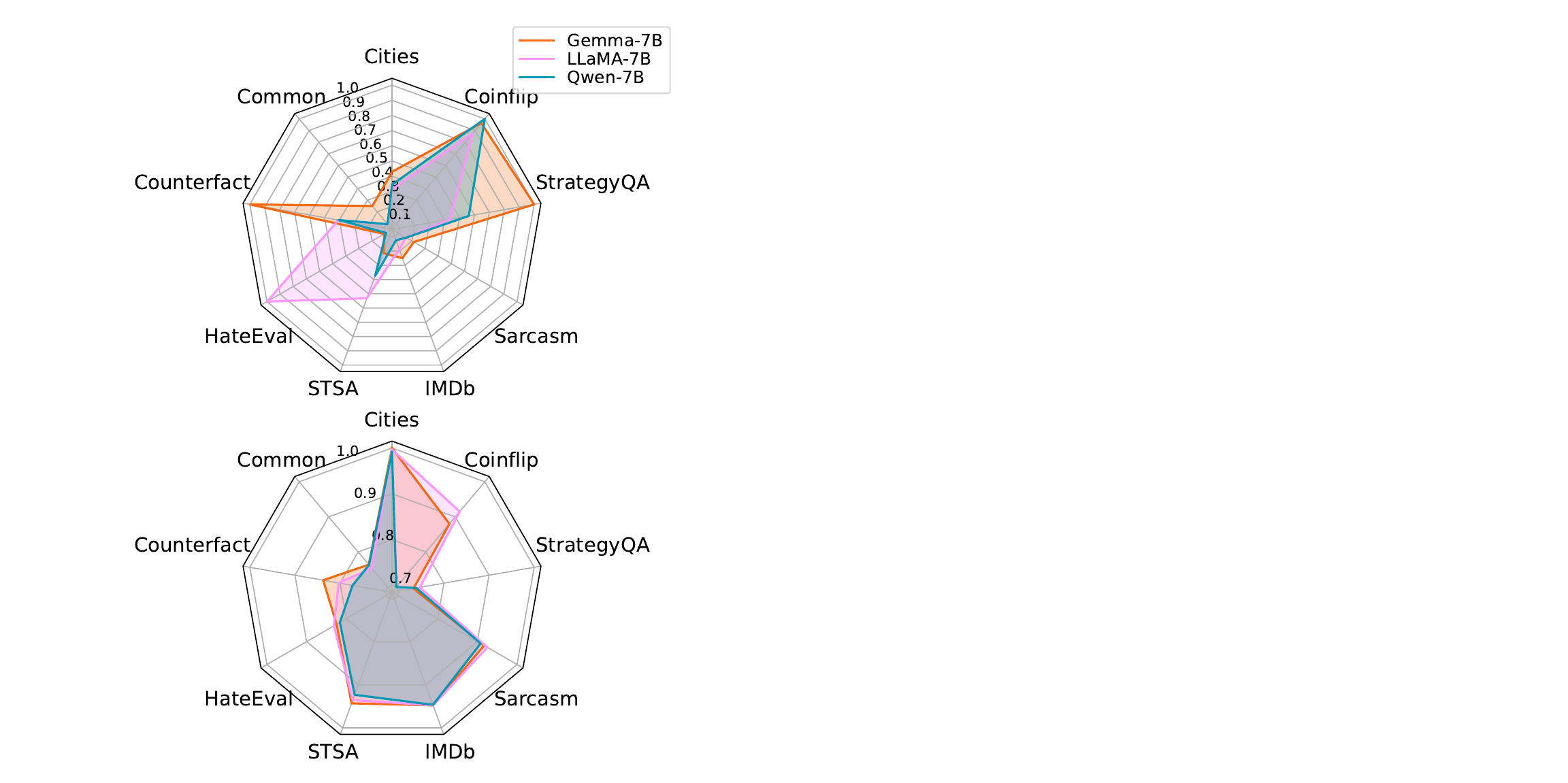}
    \caption{The upper radar image is the converging point of each dataset on Gemma-7B, LLaMA-7B, and Qwen-7B, represented by the percent depth proportion. The bottom radar image is the maximum accuracy of each dataset on Gemma-7B, LLaMA-7B, and Qwen-7B, represented by the percentage depth proportion.}
    \label{fig:5341}
    \vspace{-0.5cm}
\end{figure}

We examine how LLMs from various families, possessing a similar parameter count, process concepts as reflected by their converging points and peak accuracies. The overarching trends are highlighted in \autoref{fig:5341}, with detailed statistics on a layer-by-layer basis provided in \autoref{7b:general}. Our findings reveal that while LLMs across different families may reach nearly identical peak accuracies, the layers at which they converge to these peaks can vary. For instance, in the \textbf{HateEval} and \textbf{Counterfact} datasets, we observe models converging at significantly deeper layers. This variation suggests that despite similar parameter scales, different models may employ varied mechanisms to tackle the same problems, reflecting the diversity in how models interpret and process complex information.
\begin{tcolorbox}[colback=gray!10!white,colframe=gray!70!black,boxrule=0.3pt,title=Remark 3]
With the same number of model parameters, the models generally have a comparable understanding of the datasets.
\end{tcolorbox}

\subsection{Ablation Study}
To quantify the robustness of the LLMs concerning their internal representation, we conducted ablation studies on noise perturbation and bit quantization. The result shows that adding noises or reducing model weights to 8 bits can make the accuracy converge slower. Compressing the LLMs to 16 bits doesn't harm the understanding process too much. Details can be found in Section \ref{sec:ablastudy}.

\section{Conclusions}
This paper proposes Concept Depth, the phenomenon that different concepts are learned in different layers of LLMs, i.e., more difficult concepts are fully acquired with deeper layers. We conducted several experiments around Concept Depth using probe techniques. Our research suggests that LLMs tend to effectively categorize easy tasks, indicating that these concepts are learned in the first few initial layers. In contrast, complex tasks may only be recognizable (if at all) in deeper layers, and LLMs of the same size perform largely consistently across datasets regarding Concept Depth. Compressing the model weight to 16-bit representations for future LLMs' designs is also a promising method for saving computation memory.

\section{Limitations}

The paper presents several opportunities for further exploration. Firstly, the datasets employed might not encompass the full spectrum of language tasks, offering a chance to expand the scope of the findings in a multilingual environment. Secondly, We did not experiment with very large open-source language models, thus allowing future researchers to investigate how scaling up the model size affects concept acquisition across different layers and enhances robustness. Moreover, we should also try different kinds of classifiers, including nonlinear models and neural network-based classifiers, to acquire more profound insights into how LLM representations differ across layers. These aspects highlight promising directions for continued advancement in the field. We will continue to explore intermediate representations to help us better understand the inner side of LLMs, as this challenge may also be open to other researchers in this field.

\section*{Acknowledgement}

We thank Taowen Wang, Fei Sun, Wujiang Xu, and Guangyan Sun for their valuable discussions and suggestions during the project.

\bibliography{reference}

\begin{thebibliography}{50}
\providecommand{\natexlab}[1]{#1}

\bibitem[{Achiam et~al.(2023)Achiam, Adler, Agarwal, Ahmad, Akkaya, Aleman, Almeida, Altenschmidt, Altman, Anadkat et~al.}]{achiam2023gpt}
Josh Achiam, Steven Adler, Sandhini Agarwal, Lama Ahmad, Ilge Akkaya, Florencia~Leoni Aleman, Diogo Almeida, Janko Altenschmidt, Sam Altman, Shyamal Anadkat, et~al. 2023.
\newblock Gpt-4 technical report.
\newblock \emph{arXiv preprint arXiv:2303.08774}.

\bibitem[{Alain and Bengio(2016)}]{alain2016understanding}
Guillaume Alain and Yoshua Bengio. 2016.
\newblock Understanding intermediate layers using linear classifier probes.
\newblock \emph{arXiv preprint arXiv:1610.01644}.

\bibitem[{Armstrong et~al.(1983)Armstrong, Gleitman, and Gleitman}]{armstrong1983some}
Sharon~Lee Armstrong, Lila~R Gleitman, and Henry Gleitman. 1983.
\newblock What some concepts might not be.
\newblock \emph{Cognition}, 13(3):263--308.

\bibitem[{Azaria and Mitchell(2023)}]{azaria2023internal}
Amos Azaria and Tom Mitchell. 2023.
\newblock The internal state of an llm knows when its lying.
\newblock \emph{arXiv preprint arXiv:2304.13734}.

\bibitem[{Bai et~al.(2023)Bai, Bai, Chu, Cui, Dang, Deng, Fan, Ge, Han, Huang, Hui, Ji, Li, Lin, Lin, Liu, Liu, Lu, Lu, Ma, Men, Ren, Ren, Tan, Tan, Tu, Wang, Wang, Wang, Wu, Xu, Xu, Yang, Yang, Yang, Yang, Yao, Yu, Yuan, Yuan, Zhang, Zhang, Zhang, Zhang, Zhou, Zhou, Zhou, and Zhu}]{bai2023qwen}
Jinze Bai, Shuai Bai, Yunfei Chu, Zeyu Cui, Kai Dang, Xiaodong Deng, Yang Fan, Wenbin Ge, Yu~Han, Fei Huang, Binyuan Hui, Luo Ji, Mei Li, Junyang Lin, Runji Lin, Dayiheng Liu, Gao Liu, Chengqiang Lu, Keming Lu, Jianxin Ma, Rui Men, Xingzhang Ren, Xuancheng Ren, Chuanqi Tan, Sinan Tan, Jianhong Tu, Peng Wang, Shijie Wang, Wei Wang, Shengguang Wu, Benfeng Xu, Jin Xu, An~Yang, Hao Yang, Jian Yang, Shusheng Yang, Yang Yao, Bowen Yu, Hongyi Yuan, Zheng Yuan, Jianwei Zhang, Xingxuan Zhang, Yichang Zhang, Zhenru Zhang, Chang Zhou, Jingren Zhou, Xiaohuan Zhou, and Tianhang Zhu. 2023.
\newblock \href {https://arxiv.org/abs/2309.16609} {Qwen technical report}.
\newblock \emph{Preprint}, arXiv:2309.16609.

\bibitem[{Brown et~al.(2020)Brown, Mann, Ryder, Subbiah, Kaplan, Dhariwal, Neelakantan, Shyam, Sastry, Askell et~al.}]{brown2020language}
Tom Brown, Benjamin Mann, Nick Ryder, Melanie Subbiah, Jared~D Kaplan, Prafulla Dhariwal, Arvind Neelakantan, Pranav Shyam, Girish Sastry, Amanda Askell, et~al. 2020.
\newblock Language models are few-shot learners.
\newblock \emph{Advances in neural information processing systems}, 33:1877--1901.

\bibitem[{Casper et~al.(2023)Casper, Lin, Kwon, Culp, and Hadfield-Menell}]{casper2023explore}
Stephen Casper, Jason Lin, Joe Kwon, Gatlen Culp, and Dylan Hadfield-Menell. 2023.
\newblock \href {https://arxiv.org/abs/2306.09442} {Explore, establish, exploit: Red teaming language models from scratch}.
\newblock \emph{Preprint}, arXiv:2306.09442.

\bibitem[{Chang et~al.(2023)Chang, Wang, Wang, Wu, Yang, Zhu, Chen, Yi, Wang, Wang, Ye, Zhang, Chang, Yu, Yang, and Xie}]{chang2023survey}
Yupeng Chang, Xu~Wang, Jindong Wang, Yuan Wu, Linyi Yang, Kaijie Zhu, Hao Chen, Xiaoyuan Yi, Cunxiang Wang, Yidong Wang, Wei Ye, Yue Zhang, Yi~Chang, Philip~S. Yu, Qiang Yang, and Xing Xie. 2023.
\newblock \href {https://arxiv.org/abs/2307.03109} {A survey on evaluation of large language models}.
\newblock \emph{Preprint}, arXiv:2307.03109.

\bibitem[{Deng et~al.(2021)Deng, Ren, Zhang, and Zhang}]{deng2021discovering}
Huiqi Deng, Qihan Ren, Hao Zhang, and Quanshi Zhang. 2021.
\newblock Discovering and explaining the representation bottleneck of dnns.
\newblock \emph{arXiv preprint arXiv:2111.06236}.

\bibitem[{Dettmers et~al.(2024)Dettmers, Pagnoni, Holtzman, and Zettlemoyer}]{dettmers2024qlora}
Tim Dettmers, Artidoro Pagnoni, Ari Holtzman, and Luke Zettlemoyer. 2024.
\newblock Qlora: Efficient finetuning of quantized llms.
\newblock \emph{Advances in Neural Information Processing Systems}, 36.

\bibitem[{Duan et~al.(2024)Duan, Yang, and Tam}]{duan2024llms}
Hanyu Duan, Yi~Yang, and Kar~Yan Tam. 2024.
\newblock Do llms know about hallucination? an empirical investigation of llm's hidden states.
\newblock \emph{arXiv preprint arXiv:2402.09733}.

\bibitem[{Dubey et~al.(2024)Dubey, Jauhri, Pandey, Kadian, Al-Dahle, Letman, Mathur, Schelten, Yang, Fan et~al.}]{dubey2024llama}
Abhimanyu Dubey, Abhinav Jauhri, Abhinav Pandey, Abhishek Kadian, Ahmad Al-Dahle, Aiesha Letman, Akhil Mathur, Alan Schelten, Amy Yang, Angela Fan, et~al. 2024.
\newblock The llama 3 herd of models.
\newblock \emph{arXiv preprint arXiv:2407.21783}.

\bibitem[{Fan et~al.(2024)Fan, Jiang, Li, Meng, Han, Shang, Sun, Wang, and Wang}]{fan2024not}
Siqi Fan, Xin Jiang, Xiang Li, Xuying Meng, Peng Han, Shuo Shang, Aixin Sun, Yequan Wang, and Zhongyuan Wang. 2024.
\newblock Not all layers of llms are necessary during inference.
\newblock \emph{arXiv preprint arXiv:2403.02181}.

\bibitem[{Geva et~al.(2023)Geva, Bastings, Filippova, and Globerson}]{geva2023dissecting}
Mor Geva, Jasmijn Bastings, Katja Filippova, and Amir Globerson. 2023.
\newblock Dissecting recall of factual associations in auto-regressive language models.
\newblock \emph{arXiv preprint arXiv:2304.14767}.

\bibitem[{Geva et~al.(2021)Geva, Khashabi, Segal, Khot, Roth, and Berant}]{10.1162/tacl_a_00370}
Mor Geva, Daniel Khashabi, Elad Segal, Tushar Khot, Dan Roth, and Jonathan Berant. 2021.
\newblock \href {https://doi.org/10.1162/tacl_a_00370} {{Did Aristotle Use a Laptop? A Question Answering Benchmark with Implicit Reasoning Strategies}}.
\newblock \emph{Transactions of the Association for Computational Linguistics}, 9:346--361.

\bibitem[{Gromov et~al.(2024)Gromov, Tirumala, Shapourian, Glorioso, and Roberts}]{gromov2024unreasonable}
Andrey Gromov, Kushal Tirumala, Hassan Shapourian, Paolo Glorioso, and Daniel~A Roberts. 2024.
\newblock The unreasonable ineffectiveness of the deeper layers.
\newblock \emph{arXiv preprint arXiv:2403.17887}.

\bibitem[{Gurnee and Tegmark(2023)}]{gurnee2023language}
Wes Gurnee and Max Tegmark. 2023.
\newblock Language models represent space and time.
\newblock \emph{arXiv preprint arXiv:2310.02207}.

\bibitem[{Jin et~al.(2024{\natexlab{a}})Jin, Luo, Cheng, Wang, Hua, Tang, Wang, and Zhang}]{jin2024disentangling}
Mingyu Jin, Weidi Luo, Sitao Cheng, Xinyi Wang, Wenyue Hua, Ruixiang Tang, William~Yang Wang, and Yongfeng Zhang. 2024{\natexlab{a}}.
\newblock Disentangling memory and reasoning ability in large language models.
\newblock \emph{arXiv preprint arXiv:2411.13504}.

\bibitem[{Jin et~al.(2024{\natexlab{b}})Jin, Yu, Shu, Zhao, Hua, Meng, Zhang, and Du}]{jin2024impact}
Mingyu Jin, Qinkai Yu, Dong Shu, Haiyan Zhao, Wenyue Hua, Yanda Meng, Yongfeng Zhang, and Mengnan Du. 2024{\natexlab{b}}.
\newblock The impact of reasoning step length on large language models.
\newblock \emph{arXiv preprint arXiv:2401.04925}.

\bibitem[{Ju et~al.(2024)Ju, Sun, Du, Yuan, Ren, and Liu}]{ju2024large}
Tianjie Ju, Weiwei Sun, Wei Du, Xinwei Yuan, Zhaochun Ren, and Gongshen Liu. 2024.
\newblock How large language models encode context knowledge? a layer-wise probing study.
\newblock \emph{arXiv preprint arXiv:2402.16061}.

\bibitem[{Kim(2014)}]{kim-2014-convolutional}
Yoon Kim. 2014.
\newblock \href {https://doi.org/10.3115/v1/D14-1181} {Convolutional neural networks for sentence classification}.
\newblock In \emph{Proceedings of the 2014 Conference on Empirical Methods in Natural Language Processing ({EMNLP})}, pages 1746--1751, Doha, Qatar. Association for Computational Linguistics.

\bibitem[{Li et~al.(2024)Li, Patel, Vi{\'e}gas, Pfister, and Wattenberg}]{li2024inference}
Kenneth Li, Oam Patel, Fernanda Vi{\'e}gas, Hanspeter Pfister, and Martin Wattenberg. 2024.
\newblock Inference-time intervention: Eliciting truthful answers from a language model.
\newblock \emph{Advances in Neural Information Processing Systems}, 36.

\bibitem[{Maas et~al.(2011)Maas, Daly, Pham, Huang, Ng, and Potts}]{maas-EtAl:2011:ACL-HLT2011}
Andrew~L. Maas, Raymond~E. Daly, Peter~T. Pham, Dan Huang, Andrew~Y. Ng, and Christopher Potts. 2011.
\newblock \href {http://www.aclweb.org/anthology/P11-1015} {Learning word vectors for sentiment analysis}.
\newblock In \emph{Proceedings of the 49th Annual Meeting of the Association for Computational Linguistics: Human Language Technologies}, pages 142--150, Portland, Oregon, USA. Association for Computational Linguistics.

\bibitem[{Manolescu et~al.(2019)Manolescu, L{\"o}fflad, Mohamed~Saber, and Moradipour~Tari}]{manolescu-etal-2019-tueval}
Mihai Manolescu, Denise L{\"o}fflad, Adham~Nasser Mohamed~Saber, and Masoumeh Moradipour~Tari. 2019.
\newblock \href {https://doi.org/10.18653/v1/S19-2089} {{T}u{E}val at {S}em{E}val-2019 task 5: {LSTM} approach to hate speech detection in {E}nglish and {S}panish}.
\newblock In \emph{Proceedings of the 13th International Workshop on Semantic Evaluation}, pages 498--502, Minneapolis, Minnesota, USA. Association for Computational Linguistics.

\bibitem[{Marks and Tegmark(2023)}]{marks2023geometry}
Samuel Marks and Max Tegmark. 2023.
\newblock The geometry of truth: Emergent linear structure in large language model representations of true/false datasets.
\newblock \emph{arXiv preprint arXiv:2310.06824}.

\bibitem[{Men et~al.(2024)Men, Xu, Zhang, Wang, Lin, Lu, Han, and Chen}]{men2024shortgpt}
Xin Men, Mingyu Xu, Qingyu Zhang, Bingning Wang, Hongyu Lin, Yaojie Lu, Xianpei Han, and Weipeng Chen. 2024.
\newblock Shortgpt: Layers in large language models are more redundant than you expect.
\newblock \emph{arXiv preprint arXiv:2403.03853}.

\bibitem[{Meng et~al.(2022)Meng, Bau, Andonian, and Belinkov}]{meng2022locating}
Kevin Meng, David Bau, Alex Andonian, and Yonatan Belinkov. 2022.
\newblock Locating and editing factual associations in gpt.
\newblock \emph{Advances in Neural Information Processing Systems}, 35:17359--17372.

\bibitem[{Misra and Arora(2023)}]{misra2023Sarcasm}
Rishabh Misra and Prahal Arora. 2023.
\newblock \href {https://doi.org/10.1016/j.aiopen.2023.01.001} {Sarcasm detection using news headlines dataset}.
\newblock \emph{AI Open}, 4:13--18.

\bibitem[{OpenAI(2024)}]{gpt4o}
OpenAI. 2024.
\newblock \href {https://openai.com/index/hello-gpt-4o/} {Hello gpt-4o}.
\newblock \emph{OpenAI Blog}.

\bibitem[{Pal et~al.(2023)Pal, Sun, Yuan, Wallace, and Bau}]{pal2023future}
Koyena Pal, Jiuding Sun, Andrew Yuan, Byron~C Wallace, and David Bau. 2023.
\newblock Future lens: Anticipating subsequent tokens from a single hidden state.
\newblock \emph{arXiv preprint arXiv:2311.04897}.

\bibitem[{R{\"a}z(2023)}]{raz2023methods}
Tim R{\"a}z. 2023.
\newblock Methods for identifying emergent concepts in deep neural networks.
\newblock \emph{Patterns}, 4(6).

\bibitem[{Ren et~al.(2023)Ren, Li, Chen, Deng, and Zhang}]{ren2023defining}
Jie Ren, Mingjie Li, Qirui Chen, Huiqi Deng, and Quanshi Zhang. 2023.
\newblock Defining and quantifying the emergence of sparse concepts in dnns.
\newblock In \emph{Proceedings of the IEEE/CVF Conference on Computer Vision and Pattern Recognition}, pages 20280--20289.

\bibitem[{Su et~al.(2024{\natexlab{a}})Su, Li, Zhang, Zhu, Qu, Zhou, Bowen, Cheng et~al.}]{su2024living}
Zhaochen Su, Juntao Li, Jun Zhang, Tong Zhu, Xiaoye Qu, Pan Zhou, Yan Bowen, Yu~Cheng, et~al. 2024{\natexlab{a}}.
\newblock Living in the moment: Can large language models grasp co-temporal reasoning?
\newblock \emph{arXiv preprint arXiv:2406.09072}.

\bibitem[{Su et~al.(2022)Su, Tang, Guan, Li, Wu, and Zhang}]{su2022improving}
Zhaochen Su, Zecheng Tang, Xinyan Guan, Juntao Li, Lijun Wu, and Min Zhang. 2022.
\newblock Improving temporal generalization of pre-trained language models with lexical semantic change.
\newblock \emph{arXiv preprint arXiv:2210.17127}.

\bibitem[{Su et~al.(2024{\natexlab{b}})Su, Zhang, Qu, Zhu, Li, Sun, Li, Zhang, and Cheng}]{su2024conflictbank}
Zhaochen Su, Jun Zhang, Xiaoye Qu, Tong Zhu, Yanshu Li, Jiashuo Sun, Juntao Li, Min Zhang, and Yu~Cheng. 2024{\natexlab{b}}.
\newblock Conflictbank: A benchmark for evaluating the influence of knowledge conflicts in llm.
\newblock In \emph{Advances in neural information processing systems}.

\bibitem[{Su et~al.(2024{\natexlab{c}})Su, Zhang, Zhu, Qu, Li, Zhang, and Cheng}]{su2024timo}
Zhaochen Su, Jun Zhang, Tong Zhu, Xiaoye Qu, Juntao Li, Min Zhang, and Yu~Cheng. 2024{\natexlab{c}}.
\newblock Timo: Towards better temporal reasoning for language models.
\newblock \emph{arXiv preprint arXiv:2406.14192}.

\bibitem[{Team et~al.(2023)Team, Anil, Borgeaud, Wu, Alayrac, Yu, Soricut, Schalkwyk, Dai, Hauth et~al.}]{team2023gemini}
Gemini Team, Rohan Anil, Sebastian Borgeaud, Yonghui Wu, Jean-Baptiste Alayrac, Jiahui Yu, Radu Soricut, Johan Schalkwyk, Andrew~M Dai, Anja Hauth, et~al. 2023.
\newblock Gemini: a family of highly capable multimodal models.
\newblock \emph{arXiv preprint arXiv:2312.11805}.

\bibitem[{Team et~al.(2024)Team, Mesnard, Hardin, Dadashi, Bhupatiraju, Pathak, Sifre, Rivière, Kale, Love, Tafti, Hussenot, Chowdhery, Roberts, Barua, Botev, Castro-Ros, Slone, Héliou, Tacchetti, Bulanova, Paterson, Tsai, Shahriari, Lan, Choquette-Choo, Crepy, Cer, Ippolito, Reid, Buchatskaya, Ni, Noland, Yan, Tucker, Muraru, Rozhdestvenskiy, Michalewski, Tenney, Grishchenko, Austin, Keeling, Labanowski, Lespiau, Stanway, Brennan, Chen, Ferret, Chiu, Mao-Jones, Lee, Yu, Millican, Sjoesund, Lee, Dixon, Reid, Mikuła, Wirth, Sharman, Chinaev, Thain, Bachem, Chang, Wahltinez, Bailey, Michel, Yotov, Sessa, Chaabouni, Comanescu, Jana, Anil, McIlroy, Liu, Mullins, Smith, Borgeaud, Girgin, Douglas, Pandya, Shakeri, De, Klimenko, Hennigan, Feinberg, Stokowiec, hui Chen, Ahmed, Gong, Warkentin, Peran, Giang, Farabet, Vinyals, Dean, Kavukcuoglu, Hassabis, Ghahramani, Eck, Barral, Pereira, Collins, Joulin, Fiedel, Senter, Andreev, and Kenealy}]{gemmateam2024gemma}
Gemma Team, Thomas Mesnard, Cassidy Hardin, Robert Dadashi, Surya Bhupatiraju, Shreya Pathak, Laurent Sifre, Morgane Rivière, Mihir~Sanjay Kale, Juliette Love, Pouya Tafti, Léonard Hussenot, Aakanksha Chowdhery, Adam Roberts, Aditya Barua, Alex Botev, Alex Castro-Ros, Ambrose Slone, Amélie Héliou, Andrea Tacchetti, Anna Bulanova, Antonia Paterson, Beth Tsai, Bobak Shahriari, Charline~Le Lan, Christopher~A. Choquette-Choo, Clément Crepy, Daniel Cer, Daphne Ippolito, David Reid, Elena Buchatskaya, Eric Ni, Eric Noland, Geng Yan, George Tucker, George-Christian Muraru, Grigory Rozhdestvenskiy, Henryk Michalewski, Ian Tenney, Ivan Grishchenko, Jacob Austin, James Keeling, Jane Labanowski, Jean-Baptiste Lespiau, Jeff Stanway, Jenny Brennan, Jeremy Chen, Johan Ferret, Justin Chiu, Justin Mao-Jones, Katherine Lee, Kathy Yu, Katie Millican, Lars~Lowe Sjoesund, Lisa Lee, Lucas Dixon, Machel Reid, Maciej Mikuła, Mateo Wirth, Michael Sharman, Nikolai Chinaev, Nithum Thain, Olivier Bachem, Oscar Chang, Oscar
  Wahltinez, Paige Bailey, Paul Michel, Petko Yotov, Pier~Giuseppe Sessa, Rahma Chaabouni, Ramona Comanescu, Reena Jana, Rohan Anil, Ross McIlroy, Ruibo Liu, Ryan Mullins, Samuel~L Smith, Sebastian Borgeaud, Sertan Girgin, Sholto Douglas, Shree Pandya, Siamak Shakeri, Soham De, Ted Klimenko, Tom Hennigan, Vlad Feinberg, Wojciech Stokowiec, Yu~hui Chen, Zafarali Ahmed, Zhitao Gong, Tris Warkentin, Ludovic Peran, Minh Giang, Clément Farabet, Oriol Vinyals, Jeff Dean, Koray Kavukcuoglu, Demis Hassabis, Zoubin Ghahramani, Douglas Eck, Joelle Barral, Fernando Pereira, Eli Collins, Armand Joulin, Noah Fiedel, Evan Senter, Alek Andreev, and Kathleen Kenealy. 2024.
\newblock \href {https://arxiv.org/abs/2403.08295} {Gemma: Open models based on gemini research and technology}.
\newblock \emph{Preprint}, arXiv:2403.08295.

\bibitem[{Thoppilan et~al.(2022)Thoppilan, De~Freitas, Hall, Shazeer, Kulshreshtha, Cheng, Jin, Bos, Baker, Du et~al.}]{thoppilan2022lamda}
Romal Thoppilan, Daniel De~Freitas, Jamie Hall, Noam Shazeer, Apoorv Kulshreshtha, Heng-Tze Cheng, Alicia Jin, Taylor Bos, Leslie Baker, Yu~Du, et~al. 2022.
\newblock Lamda: Language models for dialog applications.
\newblock \emph{arXiv preprint arXiv:2201.08239}.

\bibitem[{Touvron et~al.(2023)Touvron, Martin, Stone, Albert, Almahairi, Babaei, Bashlykov, Batra, Bhargava, Bhosale et~al.}]{touvron2023llama}
Hugo Touvron, Louis Martin, Kevin Stone, Peter Albert, Amjad Almahairi, Yasmine Babaei, Nikolay Bashlykov, Soumya Batra, Prajjwal Bhargava, Shruti Bhosale, et~al. 2023.
\newblock Llama 2: Open foundation and fine-tuned chat models.
\newblock \emph{arXiv preprint arXiv:2307.09288}.

\bibitem[{Wei et~al.(2022{\natexlab{a}})Wei, Tay, Bommasani, Raffel, Zoph, Borgeaud, Yogatama, Bosma, Zhou, Metzler et~al.}]{wei2022emergent}
Jason Wei, Yi~Tay, Rishi Bommasani, Colin Raffel, Barret Zoph, Sebastian Borgeaud, Dani Yogatama, Maarten Bosma, Denny Zhou, Donald Metzler, et~al. 2022{\natexlab{a}}.
\newblock Emergent abilities of large language models.
\newblock \emph{arXiv preprint arXiv:2206.07682}.

\bibitem[{Wei et~al.(2022{\natexlab{b}})Wei, Wang, Schuurmans, Bosma, Xia, Chi, Le, Zhou et~al.}]{wei2022chain}
Jason Wei, Xuezhi Wang, Dale Schuurmans, Maarten Bosma, Fei Xia, Ed~Chi, Quoc~V Le, Denny Zhou, et~al. 2022{\natexlab{b}}.
\newblock Chain-of-thought prompting elicits reasoning in large language models.
\newblock \emph{Advances in Neural Information Processing Systems}, 35:24824--24837.

\bibitem[{Wen(2024)}]{wen2024language}
Ximing Wen. 2024.
\newblock Language model meets prototypes: Towards interpretable text classification models through prototypical networks.
\newblock \emph{arXiv preprint arXiv:2412.03761}.

\bibitem[{Wen et~al.(2024{\natexlab{a}})Wen, Tan, and Weber}]{wen2024gaprotonet}
Ximing Wen, Wenjuan Tan, and Rosina~O Weber. 2024{\natexlab{a}}.
\newblock Gaprotonet: A multi-head graph attention-based prototypical network for interpretable text classification.
\newblock \emph{arXiv preprint arXiv:2409.13312}.

\bibitem[{Wen et~al.(2024{\natexlab{b}})Wen, Weber, Sen, Hannan, Nesbit, Chan, Goffi, Morris, Hunninghake, Villalobos et~al.}]{wen2024impact}
Ximing Wen, Rosina~O Weber, Anik Sen, Darryl Hannan, Steven~C Nesbit, Vincent Chan, Alberto Goffi, Michael Morris, John~C Hunninghake, Nicholas~E Villalobos, et~al. 2024{\natexlab{b}}.
\newblock The impact of an xai-augmented approach on binary classification with scarce data.
\newblock \emph{arXiv preprint arXiv:2407.06206}.

\bibitem[{Yang et~al.(2024)Yang, Yang, Hui, Zheng, Yu, Zhou, Li, Li, Liu, Huang et~al.}]{yang2024qwen2}
An~Yang, Baosong Yang, Binyuan Hui, Bo~Zheng, Bowen Yu, Chang Zhou, Chengpeng Li, Chengyuan Li, Dayiheng Liu, Fei Huang, et~al. 2024.
\newblock Qwen2 technical report.
\newblock \emph{arXiv preprint arXiv:2407.10671}.

\bibitem[{Yeh et~al.(2019)Yeh, Kim, Arik, Li, Ravikumar, and Pfister}]{yeh2019concept}
Chih-Kuan Yeh, Been Kim, Sercan Arik, Chun-Liang Li, Pradeep Ravikumar, and Tomas Pfister. 2019.
\newblock On concept-based explanations in deep neural networks.

\bibitem[{Zhang et~al.(2023)Zhang, Shen, Yang, Ou, Yu, Zhuang et~al.}]{zhang2023pruning}
Mingyang Zhang, Chunhua Shen, Zhen Yang, Linlin Ou, Xinyi Yu, Bohan Zhuang, et~al. 2023.
\newblock Pruning meets low-rank parameter-efficient fine-tuning.
\newblock \emph{arXiv preprint arXiv:2305.18403}.

\bibitem[{Zhao et~al.(2024)Zhao, Yang, Lakkaraju, and Du}]{zhao2024opening}
Haiyan Zhao, Fan Yang, Himabindu Lakkaraju, and Mengnan Du. 2024.
\newblock Opening the black box of large language models: Two views on holistic interpretability.
\newblock \emph{arXiv preprint arXiv:2402.10688}.

\bibitem[{Zhou et~al.(2024)Zhou, Wang, Jin, Yao, Ye, Liu, Wang, Huang, and Huang}]{zhou2024mathattack}
Zihao Zhou, Qiufeng Wang, Mingyu Jin, Jie Yao, Jianan Ye, Wei Liu, Wei Wang, Xiaowei Huang, and Kaizhu Huang. 2024.
\newblock Mathattack: Attacking large language models towards math solving ability.
\newblock In \emph{Proceedings of the AAAI Conference on Artificial Intelligence}, volume~38, pages 19750--19758.

\end{thebibliography}

%\clearpage
\begin{appendices}

\clearpage
\section{Appendix}

Here, we provide our supplementary materials.

\subsection{Metrics for Parts of the Layers}

\definecolor{color_strategyqa}{RGB}{97, 221, 253}
\definecolor{color_coinflip}{RGB}{97, 221, 253}
\definecolor{color_cities}{RGB}{97, 221, 253}
\definecolor{color_common}{RGB}{97, 221, 253}
\definecolor{color_counterfact}{RGB}{97, 221, 253}
\definecolor{color_hateeval}{RGB}{97, 221, 253}
\definecolor{color_stsa}{RGB}{97, 221, 253}
\definecolor{color_imdb}{RGB}{97, 221, 253}
\definecolor{color_sarcasm}{RGB}{97, 221, 253}

\autoref{tab:stat} shows the experimental results for the accuracy, F1-score, and AUC metrics of parts of the first, 25\% depth, 50\% depth, 67\% depth, 83\% depth, and the last layer of each model over the nine datasets.

\begin{table*}[tb]
\renewcommand\arraystretch{1.6}
\centering

\resizebox{1.0\textwidth}{!}{%
\begin{tabular}{cccccccccccccccccccccccccccc}
    \toprule
Gemma-2B (18 Layers) & \multicolumn{3}{c}{StrategyQA } & \multicolumn{3}{c}{Coinflip} & \multicolumn{3}{c}{Cities}&\multicolumn{3}{c}{Common Claim}&\multicolumn{3}{c}{Counterfact}&\multicolumn{3}{c}{HateEval}&\multicolumn{3}{c}{STSA}&\multicolumn{3}{c}{IMDb}&\multicolumn{3}{c}{Sarcasm}\\
\hline
Metrics &ACC & AUC  & F1 &ACC & AUC  & F1 &ACC & AUC  & F1 &ACC & AUC  & F1 &ACC & AUC  & F1 &ACC & AUC  & F1 &ACC & AUC  & F1 &ACC & AUC  & F1 &ACC & AUC  & F1   \\
\midrule
1st-layer&\cellcolor{color_strategyqa!16}0.556&0.601&0.588&\cellcolor{color_coinflip!16}0.635&0.667&0.626&\cellcolor{color_cities!16}0.446&0.411&0.422&\cellcolor{color_common!16}0.582&0.61&0.584&\cellcolor{color_counterfact!16}0.502&0.509&0.518&\cellcolor{color_hateeval!16}0.74&0.822&0.742&\cellcolor{color_stsa!16}0.666&0.729&0.68&\cellcolor{color_imdb!16}0.722&0.788&0.725&\cellcolor{color_sarcasm!16}0.731&0.808&0.714\\
25\%-layer&\cellcolor{color_strategyqa!33}0.602&0.642&0.629&\cellcolor{color_coinflip!33}0.62&0.64&0.6&\cellcolor{color_cities!33}0.94&0.987&0.939&\cellcolor{color_common!33}0.637&0.683&0.633&\cellcolor{color_counterfact!33}0.675&0.745&0.664&\cellcolor{color_hateeval!33}0.793&0.89&0.794&\cellcolor{color_stsa!33}0.872&0.942&0.874&\cellcolor{color_imdb!33}0.884&0.953&0.885&\cellcolor{color_sarcasm!33}0.857&0.935&0.852\\
50\%-layer&\cellcolor{color_strategyqa!50}0.639&0.7&0.665&\cellcolor{color_coinflip!50}0.65&0.705&0.632&\cellcolor{color_cities!50}0.983&0.999&0.983&\cellcolor{color_common!50}0.648&0.699&0.642&\cellcolor{color_counterfact!50}0.729&0.826&0.716&\cellcolor{color_hateeval!50}0.802&0.891&0.8&\cellcolor{color_stsa!50}0.915&0.972&0.917&\cellcolor{color_imdb!50}0.941&0.982&0.941&\cellcolor{color_sarcasm!50}0.89&0.955&0.887\\
67\%-layer&\cellcolor{color_strategyqa!67}0.683&0.751&0.708&\cellcolor{color_coinflip!67}0.695&0.783&0.69&\cellcolor{color_cities!67}0.992&1.0&0.992&\cellcolor{color_common!67}0.671&0.727&0.666&\cellcolor{color_counterfact!67}0.766&0.852&0.758&\cellcolor{color_hateeval!67}0.809&0.891&0.806&\cellcolor{color_stsa!67}0.929&0.976&0.93&\cellcolor{color_imdb!67}0.936&0.984&0.936&\cellcolor{color_sarcasm!67}0.893&0.96&0.889\\
83\%-layer&\cellcolor{color_strategyqa!83}0.62&0.668&0.637&\cellcolor{color_coinflip!83}0.585&0.625&0.579&\cellcolor{color_cities!83}0.985&0.999&0.985&\cellcolor{color_common!83}0.652&0.703&0.645&\cellcolor{color_counterfact!83}0.704&0.782&0.696&\cellcolor{color_hateeval!83}0.807&0.89&0.803&\cellcolor{color_stsa!83}0.911&0.97&0.914&\cellcolor{color_imdb!83}0.926&0.977&0.926&\cellcolor{color_sarcasm!83}0.882&0.944&0.878\\
last-layer&\cellcolor{color_strategyqa!99}0.592&0.602&0.626&\cellcolor{color_coinflip!99}0.525&0.554&0.532&\cellcolor{color_cities!99}0.988&0.999&0.988&\cellcolor{color_common!99}0.647&0.697&0.644&\cellcolor{color_counterfact!99}0.685&0.752&0.679&\cellcolor{color_hateeval!99}0.803&0.892&0.801&\cellcolor{color_stsa!99}0.895&0.957&0.898&\cellcolor{color_imdb!99}0.941&0.981&0.941&\cellcolor{color_sarcasm!99}0.837&0.92&0.831\\

\bottomrule
Gemma-7B (28 Layers) & \multicolumn{3}{c}{StrategyQA } & \multicolumn{3}{c}{Coinflip} & \multicolumn{3}{c}{Cities}&\multicolumn{3}{c}{Common Claim}&\multicolumn{3}{c}{Counterfact}&\multicolumn{3}{c}{HateEval}&\multicolumn{3}{c}{STSA}&\multicolumn{3}{c}{IMDb}&\multicolumn{3}{c}{Sarcasm}\\
\toprule
Metrics &ACC & AUC  & F1 &ACC & AUC  & F1 &ACC & AUC  & F1 &ACC & AUC  & F1 &ACC & AUC  & F1 &ACC & AUC  & F1 &ACC & AUC  & F1 &ACC & AUC  & F1 &ACC & AUC  & F1  \\
\midrule

\textbf{Layer} & \cellcolor{color_strategyqa!16}0.519 & 0.565 & 0.569 & \cellcolor{color_coinflip!16}0.69 & 0.712 & 0.656 & \cellcolor{color_cities!16}0.59 & 0.444 & 0.439 & \cellcolor{color_common!16}0.737 & 0.609 & 0.053 & \cellcolor{color_counterfact!16}0.496 & 0.506 & 0.514 & \cellcolor{color_hateeval!16}0.737 & 0.817 & 0.741 & \cellcolor{color_stsa!16}0.668 & 0.737 & 0.679 & \cellcolor{color_imdb!16}0.73 & 0.811 & 0.735 & \cellcolor{color_sarcasm!16}0.723 & 0.793 & 0.713 \\
25\%-layer & \cellcolor{color_strategyqa!33}0.617 & 0.653 & 0.648 & \cellcolor{color_coinflip!33}0.605 & 0.654 & 0.591 & \cellcolor{color_cities!33}0.95 & 0.987 & 0.951 & \cellcolor{color_common!33}0.719 & 0.687 & 0.422 & \cellcolor{color_counterfact!33} 0.696 & 0.773 & 0.686 & \cellcolor{color_hateeval!33}0.795 & 0.875 & 0.794 & \cellcolor{color_stsa!33}0.89 & 0.958 & 0.893 & \cellcolor{color_imdb!33}0.9 & 0.958 & 0.899 & \cellcolor{color_sarcasm!33}0.885 & 0.953 & 0.882 \\
50\%-layer & \cellcolor{color_strategyqa!50}0.658 & 0.716 & 0.678 & \cellcolor{color_coinflip!50}0.765 & 0.844 & 0.761 & \cellcolor{color_cities!50}1.0 & 0.999 & 0.998 & \cellcolor{color_common!50}0.75 & 0.747 & 0.498 & \cellcolor{color_counterfact!50}0.835 & 0.912 & 0.829 & \cellcolor{color_hateeval!50}0.784 & 0.871 & 0.782 & \cellcolor{color_stsa!50}0.929 & 0.979 & 0.93 & \cellcolor{color_imdb!50}0.932 & 0.979 & 0.932 & \cellcolor{color_sarcasm!50}0.907 & 0.968 & 0.904 \\
67\%-layer & \cellcolor{color_strategyqa!67}0.733 & 0.809 & 0.744 & \cellcolor{color_coinflip!67}0.88 & 0.922 & 0.875 & \cellcolor{color_cities!67}1.0 & 0.998 & 0.995 & \cellcolor{color_common!67}0.756 & 0.75 & 0.515 & \cellcolor{color_counterfact!67}0.814 & 0.9 & 0.809 & \cellcolor{color_hateeval!67}0.784 & 0.868 & 0.782 & \cellcolor{color_stsa!67}0.936 & 0.981 & 0.937 & \cellcolor{color_imdb!67}0.922 & 0.979 & 0.923 & \cellcolor{color_sarcasm!67}0.913 & 0.972 & 0.91 \\
83\%-layer & \cellcolor{color_strategyqa!83}0.675 & 0.746 & 0.695 & \cellcolor{color_coinflip!83}0.785 & 0.818 & 0.786 & \cellcolor{color_cities!83}0.98 & 0.997 & 0.982 & \cellcolor{color_common!83}0.743 & 0.74 & 0.492 & \cellcolor{color_counterfact!83}0.776 & 0.867 & 0.768 & \cellcolor{color_hateeval!83}0.812 & 0.897 & 0.809 & \cellcolor{color_stsa!83}0.935 & 0.979 & 0.937 & \cellcolor{color_imdb!83}0.925 & 0.977 & 0.926 & \cellcolor{color_sarcasm!83}0.901 & 0.963 & 0.898 \\
last-layer & \cellcolor{color_strategyqa!99}0.604 & 0.666 & 0.625 & \cellcolor{color_coinflip!99}0.57 & 0.602 & 0.578 & \cellcolor{color_cities!99}0.95 & 0.996 & 0.972 & \cellcolor{color_common!99}0.759 & 0.748 & 0.481 & \cellcolor{color_counterfact!99} 0.729 & 0.808 & 0.721 & \cellcolor{color_hateeval!99}0.82 & 0.901 & 0.817 & \cellcolor{color_stsa!99}0.92 & 0.975 & 0.922 & \cellcolor{color_imdb!99}0.938 & 0.984 & 0.938 & \cellcolor{color_sarcasm!99} 0.862 & 0.932 & 0.86 \\

\bottomrule
LlaMA-7B (32 Layers) & \multicolumn{3}{c}{StrategyQA } & \multicolumn{3}{c}{Coinflip} & \multicolumn{3}{c}{Cities}&\multicolumn{3}{c}{Common Claim}&\multicolumn{3}{c}{Counterfact}&\multicolumn{3}{c}{HateEval}&\multicolumn{3}{c}{STSA}&\multicolumn{3}{c}{IMDb}&\multicolumn{3}{c}{Sarcasm}\\
\toprule
Metrics &ACC & AUC  & F1 &ACC & AUC  & F1 &ACC & AUC  & F1 &ACC & AUC  & F1 &ACC & AUC  & F1 &ACC & AUC  & F1 &ACC & AUC  & F1 &ACC & AUC  & F1 &ACC & AUC  & F1  \\
\midrule

1st-layer & \cellcolor{color_strategyqa!16}0.584 & 0.608 & 0.641 & \cellcolor{color_coinflip!16}0.545 & 0.525 & 0.674 & \cellcolor{color_cities!16}0.487 & 0.472 & 0.472 & \cellcolor{color_common!16}0.74 & 0.617 & 0.031 & \cellcolor{color_counterfact!16}0.493 & 0.491 & 0.522 & \cellcolor{color_hateeval!16}0.725 & 0.814 & 0.732 & \cellcolor{color_stsa!16}0.678 & 0.744 & 0.703 & \cellcolor{color_imdb!16}0.73 & 0.799 & 0.736 & \cellcolor{color_sarcasm!16}0.705 & 0.773 & 0.697 \\
25\%-layer & \cellcolor{color_strategyqa!33}0.657 & 0.712 & 0.688 & \cellcolor{color_coinflip!33}0.615 & 0.612 & 0.621 & \cellcolor{color_cities!33}0.93 & 0.978 & 0.929 & \cellcolor{color_common!33}0.736 & 0.699 & 0.441 & \cellcolor{color_counterfact!33}0.684 & 0.754 & 0.681 & \cellcolor{color_hateeval!33}0.806 & 0.887 & 0.792 & \cellcolor{color_stsa!33}0.513 & 0.913 & 0.676 & \cellcolor{color_imdb!33}0.914 & 0.972 & 0.914 & \cellcolor{color_sarcasm!33}0.89 & 0.961 & 0.886 \\
50\%-layer & \cellcolor{color_strategyqa!50}0.74 & 0.827 & 0.754 & \cellcolor{color_coinflip!50}0.915 & 0.977 & 0.907 & \cellcolor{color_cities!50}0.997 & 1.0 & 0.997 & \cellcolor{color_common!50}0.753 & 0.738 & 0.477 & \cellcolor{color_counterfact!50}0.797 & 0.894 & 0.793 & \cellcolor{color_hateeval!50}0.805 & 0.883 & 0.791 & \cellcolor{color_stsa!50}0.847 & 0.954 & 0.866 & \cellcolor{color_imdb!50}0.941 & 0.984 & 0.941 & \cellcolor{color_sarcasm!50}0.922 & 0.976 & 0.919 \\
67\%-layer & \cellcolor{color_strategyqa!67}0.729 & 0.805 & 0.744 & \cellcolor{color_coinflip!67}0.9 & 0.966 & 0.89 & \cellcolor{color_cities!67}0.995 & 1.0 & 0.995 & \cellcolor{color_common!67}0.744 & 0.729 & 0.466 & \cellcolor{color_counterfact!67}0.775 & 0.872 & 0.77 & \cellcolor{color_hateeval!67}0.795 & 0.88 & 0.779 & \cellcolor{color_stsa!67}0.896 & 0.957 & 0.902 & \cellcolor{color_imdb!67}0.939 & 0.985 & 0.939 & \cellcolor{color_sarcasm!67}0.92 & 0.973 & 0.918 \\
83\%-layer & \cellcolor{color_strategyqa!83}0.699 & 0.774 & 0.72 & \cellcolor{color_coinflip!83}0.88 & 0.961 & 0.871 & \cellcolor{color_cities!83}0.993 & 1.0 & 0.993 & \cellcolor{color_common!83}0.734 & 0.719 & 0.464 & \cellcolor{color_counterfact!83}0.744 & 0.832 & 0.735 & \cellcolor{color_hateeval!83}0.798 & 0.887 & 0.785 & \cellcolor{color_stsa!83}0.913 & 0.967 & 0.915 & \cellcolor{color_imdb!83}0.939 & 0.984 & 0.939 & \cellcolor{color_sarcasm!83}0.908 & 0.966 & 0.905 \\
last-layer & \cellcolor{color_strategyqa!99}0.67 & 0.744 & 0.69 & \cellcolor{color_coinflip!99}0.815 & 0.9 & 0.8 & \cellcolor{color_cities!99}0.993 & 1.0 & 0.993 & \cellcolor{color_common!99}0.743 & 0.731 & 0.464 & \cellcolor{color_counterfact!99}0.739 & 0.818 & 0.731 & \cellcolor{color_hateeval!99}0.831 & 0.914 & 0.83 & \cellcolor{color_stsa!99}0.935 & 0.984 & 0.937 & \cellcolor{color_imdb!99}0.944 & 0.987 & 0.944 & \cellcolor{color_sarcasm!99}0.895 & 0.955 & 0.892 \\

\bottomrule
LlaMA-13B (40 Layers) & \multicolumn{3}{c}{StrategyQA } & \multicolumn{3}{c}{Coinflip} & \multicolumn{3}{c}{Cities}&\multicolumn{3}{c}{Common Claim}&\multicolumn{3}{c}{Counterfact}&\multicolumn{3}{c}{HateEval}&\multicolumn{3}{c}{STSA}&\multicolumn{3}{c}{IMDb}&\multicolumn{3}{c}{Sarcasm}\\
\toprule
Metrics &ACC & AUC  & F1 &ACC & AUC  & F1 &ACC & AUC  & F1 &ACC & AUC  & F1 &ACC & AUC  & F1 &ACC & AUC  & F1 &ACC & AUC  & F1 &ACC & AUC  & F1 &ACC & AUC&F1   \\
\midrule
1st-layer & \cellcolor{color_strategyqa!16}0.567 & 0.601 & 0.628 & \cellcolor{color_coinflip!16}0.475 & 0.509 & 0.575 & \cellcolor{color_cities!16}0.481 & 0.463 & 0.47 & \cellcolor{color_common!16}0.741 & 0.62 & 0.034 & \cellcolor{color_counterfact!16}0.486 & 0.485 & 0.526 & \cellcolor{color_hateeval!16}0.719 & 0.807 & 0.727 & \cellcolor{color_stsa!16}0.697 & 0.769 & 0.709 & \cellcolor{color_imdb!16}0.732 & 0.795 & 0.736 & \cellcolor{color_sarcasm!16}0.692 & 0.756 & 0.688 \\
25\%-layer & \cellcolor{color_strategyqa!33}0.676 & 0.732 & 0.701 & \cellcolor{color_coinflip!33}0.53 & 0.59 & 0.515 & \cellcolor{color_cities!33}0.985 & 0.999 & 0.985 & \cellcolor{color_common!33}0.733 & 0.716 & 0.457 & \cellcolor{color_counterfact!33}0.763 & 0.859 & 0.758 & \cellcolor{color_hateeval!33}0.832 & 0.914 & 0.829 & \cellcolor{color_stsa!33}0.93 & 0.98 & 0.931 & \cellcolor{color_imdb!33}0.942 & 0.983 & 0.943 & \cellcolor{color_sarcasm!33}0.915 & 0.972 & 0.913 \\
50\%-layer & \cellcolor{color_strategyqa!50}0.763 & 0.844 & 0.771 & \cellcolor{color_coinflip!50}0.825 & 0.886 & 0.819 & \cellcolor{color_cities!50}0.993 & 1.0 & 0.993 & \cellcolor{color_common!50}0.758 & 0.751 & 0.515 & \cellcolor{color_counterfact!50}0.812 & 0.897 & 0.809 & \cellcolor{color_hateeval!50}0.839 & 0.92 & 0.836 & \cellcolor{color_stsa!50}0.939 & 0.984 & 0.94 & \cellcolor{color_imdb!50}0.945 & 0.984 & 0.945 & \cellcolor{color_sarcasm!50}0.936 & 0.983 & 0.934 \\
67\%-layer & \cellcolor{color_strategyqa!67}0.716 & 0.806 & 0.729 & \cellcolor{color_coinflip!67}0.795 & 0.882 & 0.794 & \cellcolor{color_cities!67}0.993 & 1.0 & 0.993 & \cellcolor{color_common!67}0.751 & 0.745 & 0.499 & \cellcolor{color_counterfact!67}0.776 & 0.866 & 0.772 & \cellcolor{color_hateeval!67}0.838 & 0.919 & 0.834 & \cellcolor{color_stsa!67}0.938 & 0.984 & 0.939 & \cellcolor{color_imdb!67}0.94 & 0.987 & 0.94 & \cellcolor{color_sarcasm!67}0.924 & 0.978 & 0.921 \\
83\%-layer & \cellcolor{color_strategyqa!83}0.71 & 0.795 & 0.719 & \cellcolor{color_coinflip!83}0.7 & 0.797 & 0.703 & \cellcolor{color_cities!83}0.99 & 1.0 & 0.99 & \cellcolor{color_common!83}0.741 & 0.731 & 0.487 & \cellcolor{color_counterfact!83}0.768 & 0.856 & 0.762 & \cellcolor{color_hateeval!83}0.832 & 0.912 & 0.829 & \cellcolor{color_stsa!83}0.937 & 0.983 & 0.938 & \cellcolor{color_imdb!83}0.941 & 0.985 & 0.942 & \cellcolor{color_sarcasm!83}0.922 & 0.974 & 0.919 \\
last-layer & \cellcolor{color_strategyqa!99}0.693 & 0.772 & 0.704 & \cellcolor{color_coinflip!99}0.645 & 0.715 & 0.664 & \cellcolor{color_cities!99}0.99 & 1.0 & 0.99 & \cellcolor{color_common!99}0.75 & 0.743 & 0.499 & \cellcolor{color_counterfact!99}0.76 & 0.841 & 0.752 & \cellcolor{color_hateeval!99}0.835 & 0.913 & 0.833 & \cellcolor{color_stsa!99}0.935 & 0.984 & 0.937 & \cellcolor{color_imdb!99}0.946 & 0.988 & 0.946 & \cellcolor{color_sarcasm!99}0.91 & 0.969 & 0.908 \\

\bottomrule
Qwen-0.5B (24 Layers) & \multicolumn{3}{c}{StrategyQA } & \multicolumn{3}{c}{Coinflip} & \multicolumn{3}{c}{Cities}&\multicolumn{3}{c}{Common Claim}&\multicolumn{3}{c}{Counterfact}&\multicolumn{3}{c}{HateEval}&\multicolumn{3}{c}{STSA}&\multicolumn{3}{c}{IMDb}&\multicolumn{3}{c}{Sarcasm}\\
\toprule
Metrics &ACC & AUC  & F1 &ACC & AUC  & F1 &ACC & AUC  & F1 &ACC & AUC  & F1 &ACC & AUC  & F1 &ACC & AUC  & F1 &ACC & AUC  & F1 &ACC & AUC  & F1 &ACC & AUC&F1   \\
\midrule

1st-layer & \cellcolor{color_strategyqa!16}0.557 & 0.578 & 0.607 & \cellcolor{color_coinflip!16}0.535 & 0.649 & 0.657 & \cellcolor{color_cities!16}0.482 & 0.46 & 0.464 & \cellcolor{color_common!16}0.735 & 0.622 & 0.11 & \cellcolor{color_counterfact!16}0.499 & 0.503 & 0.527 & \cellcolor{color_hateeval!16}0.759 & 0.851 & 0.764 & \cellcolor{color_stsa!16}0.73 & 0.801 & 0.733 & \cellcolor{color_imdb!16}0.764 & 0.837 & 0.762 & \cellcolor{color_sarcasm!16}0.729 & 0.799 & 0.72 \\
25\%-layer & \cellcolor{color_strategyqa!33}0.583 & 0.63 & 0.62 & \cellcolor{color_coinflip!33}0.545 & 0.582 & 0.508 & \cellcolor{color_cities!33}0.731 & 0.797 & 0.722 & \cellcolor{color_common!33}0.732 & 0.651 & 0.257 & \cellcolor{color_counterfact!33}0.52 & 0.524 & 0.523 & \cellcolor{color_hateeval!33}0.785 & 0.864 & 0.783 & \cellcolor{color_stsa!33}0.751 & 0.829 & 0.756 & \cellcolor{color_imdb!33}0.804 & 0.887 & 0.806 & \cellcolor{color_sarcasm!33}0.811 & 0.895 & 0.799 \\
50\%-layer & \cellcolor{color_strategyqa!50}0.619 & 0.686 & 0.649 & \cellcolor{color_coinflip!50}0.62 & 0.652 & 0.596 & \cellcolor{color_cities!50}0.935 & 0.979 & 0.935 & \cellcolor{color_common!50}0.744 & 0.695 & 0.379 & \cellcolor{color_counterfact!50}0.68 & 0.754 & 0.676 & \cellcolor{color_hateeval!50}0.793 & 0.88 & 0.792 & \cellcolor{color_stsa!50}0.846 & 0.921 & 0.848 & \cellcolor{color_imdb!50}0.884 & 0.949 & 0.883 & \cellcolor{color_sarcasm!50}0.838 & 0.92 & 0.831 \\
67\%-layer & \cellcolor{color_strategyqa!67}0.644 & 0.688 & 0.673 & \cellcolor{color_coinflip!67}0.585 & 0.617 & 0.561 & \cellcolor{color_cities!67}0.933 & 0.982 & 0.934 & \cellcolor{color_common!67}0.742 & 0.705 & 0.375 & \cellcolor{color_counterfact!67}0.668 & 0.74 & 0.665 & \cellcolor{color_hateeval!67}0.789 & 0.874 & 0.786 & \cellcolor{color_stsa!67}0.868 & 0.946 & 0.87 & \cellcolor{color_imdb!67}0.894 & 0.956 & 0.893 & \cellcolor{color_sarcasm!67}0.827 & 0.911 & 0.821 \\
83\%-layer & \cellcolor{color_strategyqa!83}0.583 & 0.61 & 0.612 & \cellcolor{color_coinflip!83}0.62 & 0.668 & 0.612 & \cellcolor{color_cities!83}0.923 & 0.971 & 0.924 & \cellcolor{color_common!83}0.746 & 0.706 & 0.364 & \cellcolor{color_counterfact!83}0.604 & 0.657 & 0.6 & \cellcolor{color_hateeval!83}0.791 & 0.867 & 0.791 & \cellcolor{color_stsa!83}0.85 & 0.927 & 0.854 & \cellcolor{color_imdb!83}0.866 & 0.941 & 0.865 & \cellcolor{color_sarcasm!83}0.824 & 0.902 & 0.819 \\
last-layer & \cellcolor{color_strategyqa!99}0.55 & 0.567 & 0.584 & \cellcolor{color_coinflip!99}0.55 & 0.613 & 0.541 & \cellcolor{color_cities!99}0.912 & 0.971 & 0.912 & \cellcolor{color_common!99}0.742 & 0.703 & 0.357 & \cellcolor{color_counterfact!99}0.579 & 0.616 & 0.579 & \cellcolor{color_hateeval!99}0.784 & 0.866 & 0.781 & \cellcolor{color_stsa!99}0.844 & 0.922 & 0.848 & \cellcolor{color_imdb!99}0.879 & 0.951 & 0.88 & \cellcolor{color_sarcasm!99}0.825 & 0.9 & 0.82 \\

\bottomrule
Qwen-1.8B (24 Layers) & \multicolumn{3}{c}{StrategyQA } & \multicolumn{3}{c}{Coinflip} & \multicolumn{3}{c}{Cities}&\multicolumn{3}{c}{Common Claim}&\multicolumn{3}{c}{Counterfact}&\multicolumn{3}{c}{HateEval}&\multicolumn{3}{c}{STSA}&\multicolumn{3}{c}{IMDb}&\multicolumn{3}{c}{Sarcasm}\\
\toprule
Metrics &ACC & AUC  & F1 &ACC & AUC  & F1 &ACC & AUC  & F1 &ACC & AUC  & F1 &ACC & AUC  & F1 &ACC & AUC  & F1 &ACC & AUC  & F1 &ACC & AUC  & F1 &ACC & AUC&F1   \\
\midrule

1st-layer & \cellcolor{color_strategyqa!16}0.57 & 0.6 & 0.63 & \cellcolor{color_coinflip!16}0.49 & 0.634 & 0.648 & \cellcolor{color_cities!16}0.482 & 0.458 & 0.464 & \cellcolor{color_common!16}0.739 & 0.619 & 0.071 & \cellcolor{color_counterfact!16}0.516 & 0.514 & 0.539 & \cellcolor{color_hateeval!16}0.724 & 0.819 & 0.732 & \cellcolor{color_stsa!16}0.693 & 0.762 & 0.703 & \cellcolor{color_imdb!16}0.718 & 0.784 & 0.72 & \cellcolor{color_sarcasm!16}0.721 & 0.796 & 0.713 \\
25\%-layer & \cellcolor{color_strategyqa!33}0.607 & 0.638 & 0.643 & \cellcolor{color_coinflip!33}0.58 & 0.59 & 0.584 & \cellcolor{color_cities!33}0.583 & 0.626 & 0.582 & \cellcolor{color_common!33}0.736 & 0.658 & 0.317 & \cellcolor{color_counterfact!33}0.521 & 0.541 & 0.525 & \cellcolor{color_hateeval!33}0.809 & 0.882 & 0.807 & \cellcolor{color_stsa!33}0.775 & 0.844 & 0.781 & \cellcolor{color_imdb!33}0.81 & 0.899 & 0.81 & \cellcolor{color_sarcasm!33}0.833 & 0.909 & 0.825 \\
50\%-layer & \cellcolor{color_strategyqa!50}0.658 & 0.726 & 0.676 & \cellcolor{color_coinflip!50}0.595 & 0.655 & 0.58 & \cellcolor{color_cities!50}0.975 & 0.997 & 0.975 & \cellcolor{color_common!50}0.741 & 0.708 & 0.419 & \cellcolor{color_counterfact!50}0.688 & 0.767 & 0.683 & \cellcolor{color_hateeval!50}0.808 & 0.89 & 0.807 & \cellcolor{color_stsa!50}0.895 & 0.961 & 0.897 & \cellcolor{color_imdb!50}0.914 & 0.974 & 0.915 & \cellcolor{color_sarcasm!50}0.87 & 0.947 & 0.864 \\
67\%-layer & \cellcolor{color_strategyqa!67}0.664 & 0.733 & 0.685 & \cellcolor{color_coinflip!67}0.695 & 0.759 & 0.655 & \cellcolor{color_cities!67}0.977 & 0.996 & 0.977 & \cellcolor{color_common!67}0.741 & 0.717 & 0.423 & \cellcolor{color_counterfact!67}0.695 & 0.776 & 0.689 & \cellcolor{color_hateeval!67}0.809 & 0.886 & 0.804 & \cellcolor{color_stsa!67}0.893 & 0.963 & 0.895 & \cellcolor{color_imdb!67}0.904 & 0.968 & 0.905 & \cellcolor{color_sarcasm!67}0.865 & 0.943 & 0.859 \\
83\%-layer & \cellcolor{color_strategyqa!83}0.631 & 0.666 & 0.649 & \cellcolor{color_coinflip!83}0.7 & 0.747 & 0.674 & \cellcolor{color_cities!83}0.972 & 0.995 & 0.972 & \cellcolor{color_common!83}0.735 & 0.706 & 0.419 & \cellcolor{color_counterfact!83}0.657 & 0.734 & 0.651 & \cellcolor{color_hateeval!83}0.791 & 0.877 & 0.788 & \cellcolor{color_stsa!83}0.89 & 0.956 & 0.893 & \cellcolor{color_imdb!83}0.891 & 0.957 & 0.893 & \cellcolor{color_sarcasm!83}0.835 & 0.922 & 0.829 \\
last-layer & \cellcolor{color_strategyqa!99}0.595 & 0.642 & 0.604 & \cellcolor{color_coinflip!99}0.615 & 0.667 & 0.605 & \cellcolor{color_cities!99}0.973 & 0.996 & 0.973 & \cellcolor{color_common!99}0.74 & 0.704 & 0.404 & \cellcolor{color_counterfact!99}0.638 & 0.713 & 0.631 & \cellcolor{color_hateeval!99}0.795 & 0.877 & 0.794 & \cellcolor{color_stsa!99}0.879 & 0.949 & 0.882 & \cellcolor{color_imdb!99}0.906 & 0.962 & 0.907 & \cellcolor{color_sarcasm!99}0.812 & 0.896 & 0.808 \\

\bottomrule
Qwen-7B (40 Layers) & \multicolumn{3}{c}{StrategyQA } & \multicolumn{3}{c}{Coinflip} & \multicolumn{3}{c}{Cities}&\multicolumn{3}{c}{Common Claim}&\multicolumn{3}{c}{Counterfact}&\multicolumn{3}{c}{HateEval}&\multicolumn{3}{c}{STSA}&\multicolumn{3}{c}{IMDb}&\multicolumn{3}{c}{Sarcasm}\\
\toprule
Metrics &ACC & AUC  & F1 &ACC & AUC  & F1 &ACC & AUC  & F1 &ACC & AUC  & F1 &ACC & AUC  & F1 &ACC & AUC  & F1 &ACC & AUC  & F1 &ACC & AUC  & F1 &ACC & AUC&F1   \\
\midrule

1st-layer & \cellcolor{color_strategyqa!16}0.54 & 0.559 & 0.602 & \cellcolor{color_coinflip!16}0.545 & 0.597 & 0.588 & \cellcolor{color_cities!16}0.437 & 0.403 & 0.418 & \cellcolor{color_common!16}0.735 & 0.635 & 0.169 & \cellcolor{color_counterfact!16}0.475 & 0.466 & 0.498 & \cellcolor{color_hateeval!16}0.785 & 0.867 & 0.786 & \cellcolor{color_stsa!16}0.753 & 0.826 & 0.757 & \cellcolor{color_imdb!16}0.782 & 0.858 & 0.785 & \cellcolor{color_sarcasm!16}0.771 & 0.856 & 0.765 \\
25\%-layer & \cellcolor{color_strategyqa!33}0.59 & 0.625 & 0.631 & \cellcolor{color_coinflip!33}0.62 & 0.635 & 0.631 & \cellcolor{color_cities!33}0.806 & 0.89 & 0.803 & \cellcolor{color_common!33}0.721 & 0.657 & 0.348 & \cellcolor{color_counterfact!33}0.556 & 0.57 & 0.553 & \cellcolor{color_hateeval!33}0.798 & 0.879 & 0.795 & \cellcolor{color_stsa!33}0.782 & 0.869 & 0.787 & \cellcolor{color_imdb!33}0.825 & 0.916 & 0.827 & \cellcolor{color_sarcasm!33}0.852 & 0.924 & 0.848 \\
50\%-layer & \cellcolor{color_strategyqa!50}0.705 & 0.782 & 0.724 & \cellcolor{color_coinflip!50}0.66 & 0.719 & 0.667 & \cellcolor{color_cities!50}0.985 & 0.998 & 0.985 & \cellcolor{color_common!50}0.744 & 0.733 & 0.452 & \cellcolor{color_counterfact!50}0.731 & 0.825 & 0.722 & \cellcolor{color_hateeval!50}0.802 & 0.885 & 0.795 & \cellcolor{color_stsa!50}0.912 & 0.969 & 0.914 & \cellcolor{color_imdb!50}0.929 & 0.975 & 0.929 & \cellcolor{color_sarcasm!50}0.882 & 0.956 & 0.878 \\
67\%-layer & \cellcolor{color_strategyqa!67}0.702 & 0.773 & 0.722 & \cellcolor{color_coinflip!67}0.635 & 0.719 & 0.64 & \cellcolor{color_cities!67}0.983 & 0.997 & 0.983 & \cellcolor{color_common!67}0.746 & 0.728 & 0.465 & \cellcolor{color_counterfact!67}0.721 & 0.801 & 0.716 & \cellcolor{color_hateeval!67}0.793 & 0.88 & 0.789 & \cellcolor{color_stsa!67}0.904 & 0.965 & 0.907 & \cellcolor{color_imdb!67}0.915 & 0.968 & 0.913 & \cellcolor{color_sarcasm!67}0.89 & 0.955 & 0.888 \\
83\%-layer & \cellcolor{color_strategyqa!83}0.678 & 0.751 & 0.703 & \cellcolor{color_coinflip!83}0.685 & 0.762 & 0.693 & \cellcolor{color_cities!83}0.978 & 0.996 & 0.978 & \cellcolor{color_common!83}0.747 & 0.724 & 0.45 & \cellcolor{color_counterfact!83}0.68 & 0.747 & 0.667 & \cellcolor{color_hateeval!83}0.795 & 0.881 & 0.79 & \cellcolor{color_stsa!83}0.899 & 0.961 & 0.901 & \cellcolor{color_imdb!83}0.916 & 0.965 & 0.915 & \cellcolor{color_sarcasm!83}0.872 & 0.941 & 0.868 \\
last-layer & \cellcolor{color_strategyqa!99}0.676 & 0.718 & 0.693 & \cellcolor{color_coinflip!99}0.585 & 0.623 & 0.587 & \cellcolor{color_cities!99}0.982 & 0.995 & 0.981 & \cellcolor{color_common!99}0.75 & 0.727 & 0.453 & \cellcolor{color_counterfact!99}0.657 & 0.718 & 0.643 & \cellcolor{color_hateeval!99}0.782 & 0.869 & 0.778 & \cellcolor{color_stsa!99}0.902 & 0.964 & 0.905 & \cellcolor{color_imdb!99}0.92 & 0.973 & 0.919 & \cellcolor{color_sarcasm!99}0.848 & 0.921 & 0.845 \\

\bottomrule
Qwen-14B (32 Layers) & \multicolumn{3}{c}{StrategyQA } & \multicolumn{3}{c}{Coinflip} & \multicolumn{3}{c}{Cities}&\multicolumn{3}{c}{Common Claim}&\multicolumn{3}{c}{Counterfact}&\multicolumn{3}{c}{HateEval}&\multicolumn{3}{c}{STSA}&\multicolumn{3}{c}{IMDb}&\multicolumn{3}{c}{Sarcasm}\\
\toprule
Metrics &ACC & AUC  & F1 &ACC & AUC  & F1 &ACC & AUC  & F1 &ACC & AUC  & F1 &ACC & AUC  & F1 &ACC & AUC  & F1 &ACC & AUC  & F1 &ACC & AUC  & F1 &ACC & AUC&F1   \\
\midrule

1st-layer & \cellcolor{color_strategyqa!16}0.582 & 0.608 & 0.624 & \cellcolor{color_coinflip!16}0.555 & 0.599 & 0.594 & \cellcolor{color_cities!16}0.436 & 0.399 & 0.429 & \cellcolor{color_common!16}0.738 & 0.627 & 0.205 & \cellcolor{color_counterfact!16}0.487 & 0.481 & 0.491 & \cellcolor{color_hateeval!16}0.77 & 0.864 & 0.772 & \cellcolor{color_stsa!16}0.749 & 0.827 & 0.754 & \cellcolor{color_imdb!16}0.768 & 0.864 & 0.768 & \cellcolor{color_sarcasm!16}0.785 & 0.866 & 0.773 \\
25\%-layer & \cellcolor{color_strategyqa!33}0.605 & 0.645 & 0.644 & \cellcolor{color_coinflip!33}0.555 & 0.609 & 0.566 & \cellcolor{color_cities!33}0.858 & 0.93 & 0.855 & \cellcolor{color_common!33}0.713 & 0.667 & 0.362 & \cellcolor{color_counterfact!33}0.558 & 0.601 & 0.555 & \cellcolor{color_hateeval!33}0.8 & 0.879 & 0.796 & \cellcolor{color_stsa!33}0.818 & 0.898 & 0.821 & \cellcolor{color_imdb!33}0.861 & 0.936 & 0.864 & \cellcolor{color_sarcasm!33}0.876 & 0.946 & 0.871 \\
50\%-layer & \cellcolor{color_strategyqa!50}0.695 & 0.775 & 0.709 & \cellcolor{color_coinflip!50}0.69 & 0.731 & 0.69 & \cellcolor{color_cities!50}0.99 & 0.999 & 0.99 & \cellcolor{color_common!50}0.744 & 0.742 & 0.483 & \cellcolor{color_counterfact!50}0.771 & 0.861 & 0.762 & \cellcolor{color_hateeval!50}0.812 & 0.897 & 0.807 & \cellcolor{color_stsa!50}0.918 & 0.972 & 0.919 & \cellcolor{color_imdb!50}0.946 & 0.983 & 0.946 & \cellcolor{color_sarcasm!50}0.901 & 0.966 & 0.897 \\
67\%-layer & \cellcolor{color_strategyqa!67}0.734 & 0.82 & 0.749 & \cellcolor{color_coinflip!67}0.685 & 0.753 & 0.693 & \cellcolor{color_cities!67}0.992 & 0.997 & 0.992 & \cellcolor{color_common!67}0.762 & 0.754 & 0.5 & \cellcolor{color_counterfact!67}0.769 & 0.848 & 0.765 & \cellcolor{color_hateeval!67}0.802 & 0.889 & 0.798 & \cellcolor{color_stsa!67}0.914 & 0.974 & 0.916 & \cellcolor{color_imdb!67}0.93 & 0.978 & 0.93 & \cellcolor{color_sarcasm!67}0.898 & 0.963 & 0.895 \\
83\%-layer & \cellcolor{color_strategyqa!83}0.737 & 0.814 & 0.751 & \cellcolor{color_coinflip!83}0.68 & 0.74 & 0.677 & \cellcolor{color_cities!83}0.992 & 0.996 & 0.992 & \cellcolor{color_common!83}0.751 & 0.732 & 0.485 & \cellcolor{color_counterfact!83}0.726 & 0.808 & 0.719 & \cellcolor{color_hateeval!83}0.806 & 0.886 & 0.805 & \cellcolor{color_stsa!83}0.91 & 0.972 & 0.912 & \cellcolor{color_imdb!83}0.925 & 0.974 & 0.925 & \cellcolor{color_sarcasm!83}0.892 & 0.957 & 0.887 \\
last-layer & \cellcolor{color_strategyqa!99}0.714 & 0.785 & 0.729 & \cellcolor{color_coinflip!99}0.65 & 0.681 & 0.646 & \cellcolor{color_cities!99}0.982 & 0.996 & 0.981 & \cellcolor{color_common!99}0.753 & 0.74 & 0.483 & \cellcolor{color_counterfact!99}0.707 & 0.779 & 0.701 & \cellcolor{color_hateeval!99}0.802 & 0.881 & 0.798 & \cellcolor{color_stsa!99}0.908 & 0.97 & 0.91 & \cellcolor{color_imdb!99}0.932 & 0.979 & 0.933 & \cellcolor{color_sarcasm!99}0.871 & 0.943 & 0.867 \\

\bottomrule
Qwen-14B (40 Layers) & \multicolumn{3}{c}{StrategyQA } & \multicolumn{3}{c}{Coinflip} & \multicolumn{3}{c}{Cities}&\multicolumn{3}{c}{Common Claim}&\multicolumn{3}{c}{Counterfact}&\multicolumn{3}{c}{HateEval}&\multicolumn{3}{c}{STSA}&\multicolumn{3}{c}{IMDb}&\multicolumn{3}{c}{Sarcasm}\\
\toprule
Metrics &ACC & AUC  & F1 &ACC & AUC  & F1 &ACC & AUC  & F1 &ACC & AUC  & F1 &ACC & AUC  & F1 &ACC & AUC  & F1 &ACC & AUC  & F1 &ACC & AUC  & F1 &ACC & AUC&F1   \\
\midrule
1st-layer & \cellcolor{color_strategyqa!16}0.569 & 0.59 & 0.626 & \cellcolor{color_coinflip!16}0.605 & 0.639 & 0.646 & \cellcolor{color_cities!16}0.461 & 0.422 & 0.448 & \cellcolor{color_common!16}0.733 & 0.62 & 0.162 & \cellcolor{color_counterfact!16}0.481 & 0.486 & 0.504 & \cellcolor{color_hateeval!16}0.766 & 0.86 & 0.769 & \cellcolor{color_stsa!16}0.75 & 0.824 & 0.753 & \cellcolor{color_imdb!16}0.778 & 0.856 & 0.776 & \cellcolor{color_sarcasm!16}0.772 & 0.859 & 0.762 \\
25\%-layer & \cellcolor{color_strategyqa!33}0.603 & 0.637 & 0.631 & \cellcolor{color_coinflip!33}0.68 & 0.704 & 0.66 & \cellcolor{color_cities!33}0.866 & 0.945 & 0.865 & \cellcolor{color_common!33}0.72 & 0.669 & 0.393 & \cellcolor{color_counterfact!33}0.621 & 0.67 & 0.615 & \cellcolor{color_hateeval!33}0.8 & 0.89 & 0.797 & \cellcolor{color_stsa!33}0.842 & 0.919 & 0.845 & \cellcolor{color_imdb!33}0.862 & 0.939 & 0.861 & \cellcolor{color_sarcasm!33}0.874 & 0.946 & 0.869 \\
50\%-layer & \cellcolor{color_strategyqa!50}0.735 & 0.835 & 0.747 & \cellcolor{color_coinflip!50}0.71 & 0.754 & 0.681 & \cellcolor{color_cities!50}0.995 & 1.0 & 0.995 & \cellcolor{color_common!50}0.76 & 0.748 & 0.515 & \cellcolor{color_counterfact!50}0.791 & 0.875 & 0.787 & \cellcolor{color_hateeval!50}0.808 & 0.891 & 0.807 & \cellcolor{color_stsa!50}0.932 & 0.981 & 0.934 & \cellcolor{color_imdb!50}0.935 & 0.984 & 0.935 & \cellcolor{color_sarcasm!50}0.919 & 0.974 & 0.916 \\
67\%-layer & \cellcolor{color_strategyqa!67}0.798 & 0.883 & 0.808 & \cellcolor{color_coinflip!67}0.7 & 0.791 & 0.674 & \cellcolor{color_cities!67}0.995 & 1.0 & 0.995 & \cellcolor{color_common!67}0.761 & 0.76 & 0.5 & \cellcolor{color_counterfact!67}0.775 & 0.859 & 0.769 & \cellcolor{color_hateeval!67}0.816 & 0.894 & 0.814 & \cellcolor{color_stsa!67}0.933 & 0.98 & 0.934 & \cellcolor{color_imdb!67}0.93 & 0.978 & 0.93 & \cellcolor{color_sarcasm!67}0.914 & 0.968 & 0.911 \\
83\%-layer & \cellcolor{color_strategyqa!83}0.796 & 0.872 & 0.807 & \cellcolor{color_coinflip!83}0.715 & 0.768 & 0.705 & \cellcolor{color_cities!83}0.992 & 1.0 & 0.992 & \cellcolor{color_common!83}0.753 & 0.735 & 0.493 & \cellcolor{color_counterfact!83}0.753 & 0.843 & 0.748 & \cellcolor{color_hateeval!83}0.816 & 0.896 & 0.811 & \cellcolor{color_stsa!83}0.927 & 0.978 & 0.927 & \cellcolor{color_imdb!83}0.936 & 0.978 & 0.936 & \cellcolor{color_sarcasm!83}0.912 & 0.964 & 0.91 \\
last-layer & \cellcolor{color_strategyqa!99}0.783 & 0.863 & 0.792 & \cellcolor{color_coinflip!99}0.605 & 0.67 & 0.586 & \cellcolor{color_cities!99}0.99 & 1.0 & 0.99 & \cellcolor{color_common!99}0.757 & 0.749 & 0.481 & \cellcolor{color_counterfact!99}0.725 & 0.814 & 0.723 & \cellcolor{color_hateeval!99}0.815 & 0.897 & 0.812 & \cellcolor{color_stsa!99}0.922 & 0.978 & 0.923 & \cellcolor{color_imdb!99}0.931 & 0.978 & 0.93 & \cellcolor{color_sarcasm!99}0.89 & 0.952 & 0.887 \\

    \bottomrule

\end{tabular}%
}
\caption{The experimental results of the nine datasets on nine LLMs. For each LLM, we select six layers (first, 25\% depth, 50\% depth, 67\% depth, 83\% depth, last) to record the accuracy, F1-score, and AUC.}
\label{tab:stat}

\end{table*}
\subsection{Description of the dataset}
\label{adata}
\noindent \textbf{Cities} \cite{marks2023geometry}:  consists of statements about the location of cities and their veracity labels (e.g., The city of Zagreb is in Japan, which is wrong). We use 1496 of these samples.

\noindent \textbf{CommonClaim} \cite{casper2023explore}: A dataset of boolean statements, each labeled by two humans as common-knowledge-true, common-knowledge-false, or neither. We use 6000 of these samples.

\noindent \textbf{Counterfact} \cite{meng2022locating}: Counterfact includes myriad counterfactuals that allows quantitative testing of specificity and generalization when learning a counterfactual. We use 4000 of these samples.

% , which are in similar formats, consist of around 50,000 commonsense statements and their veracity labels (e.g., Belgium is known for its brassieres, which is right)
%\noindent
%\textbf{High Level Tasks}
%\noindent
%We categorized the datasets into two groups based on their focus: one on emotional comprehension and the other on logical deduction, representing tasks that require advanced levels of understanding. Achieving accuracy in these tasks requires a profound interpretation of the content. For emotion task datasets, we used:
\noindent  \textbf{HateEval} \cite{manolescu-etal-2019-tueval}: HateEval has English tweets which were annotated hierarchically.  We use 6000 of these samples.

% We use the hate speech annotations (i.e., hate} or non-hate}) within this dataset as the labels of interest in our study. 

\noindent \textbf{STSA} \cite{kim-2014-convolutional}: STSA includes movie reviews, half of which were considered positive and the other half negative. Each label is extracted from a longer movie review and reflects the writer's overall intention for this review. We use 6920 of these samples.

\noindent \textbf{IMDb} \cite{maas-EtAl:2011:ACL-HLT2011}: IMDb is a benchmark dataset for binary sentiment classification. We use 2000 of these samples.

% and \textbf{IMDb} \cite{maas-EtAl:2011:ACL-HLT2011}: STSA and IMDb are sentiment analysis datasets which are annotated as positive or negative. They have 9,613 and 50,000 entries, respectively.

% \textbf{IMDb} \cite{maas-EtAl:2011:ACL-HLT2011}: A benchmark dataset for sentiment analysis. It contains 50,000 movie reviews expressing two diametrically opposed attitudes.

\noindent \textbf{Sarcasm} \cite{misra2023Sarcasm}: Sarcasm is a superior news headline dataset that tells if the headlines are sarcastic. We use 6000 of these samples.

% For logical reasoning datasets, we used:

\noindent \textbf{StrategyQA} \cite{10.1162/tacl_a_00370}: StrategyQA contains questions across all knowledge domains to elicit creative and diverse yes/no questions that require implicit reasoning steps. We use 2290 of these samples.

% StrategyQA contains 2,780 multi-hop reasoning problems which require well-established reasoning abilities to reach the correct decision. For instance, for a problem like \textit{Would a pear sink in water?}, the model has to understand and reason over the density to output correctly.

\noindent \textbf{Coinflip} \cite{wei2022chain}: Coinflip includes coin flipping queries, asking if a coin remains heads up after it is either flipped or left unflipped by individuals. We use 500 of these samples.

\subsection{Ablation Study}

\label{sec:ablastudy}
% \subsubsection{Adding Noise}
\textbf{Adding Noise.} To quantify the robustness of the LLMs concerning their internal representation, when we input these questions into the LLM, we add a random string of noise in front of the question to interfere. For instance, we perturbate a question $q$ into $q' = N \oplus q$. Here, $\oplus$ is the concatenate operation, and the noise string $N\in \{S_1,S_2\}$ disrupts the classification tasks, satisfying $$P(N=S_1)=P(N=S_2)=50\%$$
The probability distribution of $N$ is unrelated to their labels. The following is an example of \textbf{STSA} with a positive label.

\begin{tcolorbox}[colback=white!95!gray, colframe=black, width=0.5\textwidth, arc=4mm, boxrule=0.5mm]
\noindent \textbf{Before adding noise:}\\ \texttt{\color{blue} The production values are of the highest and the performances attractive without being memorable.} \texttt{The sentence above is a movie review and reflects the writer's overall intention for this review. According to the sentence, judge whether the emotion is Positive or Negative.}

\vspace{0.1cm}

\noindent \textbf{After adding noise:}\\ \texttt{\color{red} aaa} \texttt{\color{blue} The production values are of the highest and the performances attractive without being memorable.} \texttt{The sentence above is a movie review and reflects the writer's overall intention for this review. According to the sentence, judge whether the emotion is Positive or Negative.}
\\
\\
\noindent \texttt{\color{red} aaa} can be substituted by \texttt{\color{red} bbb}, both cases have equal probability.

\end{tcolorbox}

% \subsubsection{Quantization Settings}
\noindent \textbf{Quantization Settings.} Quantization has been a rule of thumb for faster LLMs' inference. The general fact is that using a lower precision level allows LLMs to be run on less capable hardware with an acceptable reduction in their ability and accuracy. Our approach to applying quantization is to explore its effect on Concept Depth, whether it still maintains the previous Concept Depth or slows down the process of understanding knowledge. We quantize the model weights to 8, 16, and 32 bits.
 
\begin{figure}[t]
\begin{center}
    \includegraphics[width=0.4\textwidth]{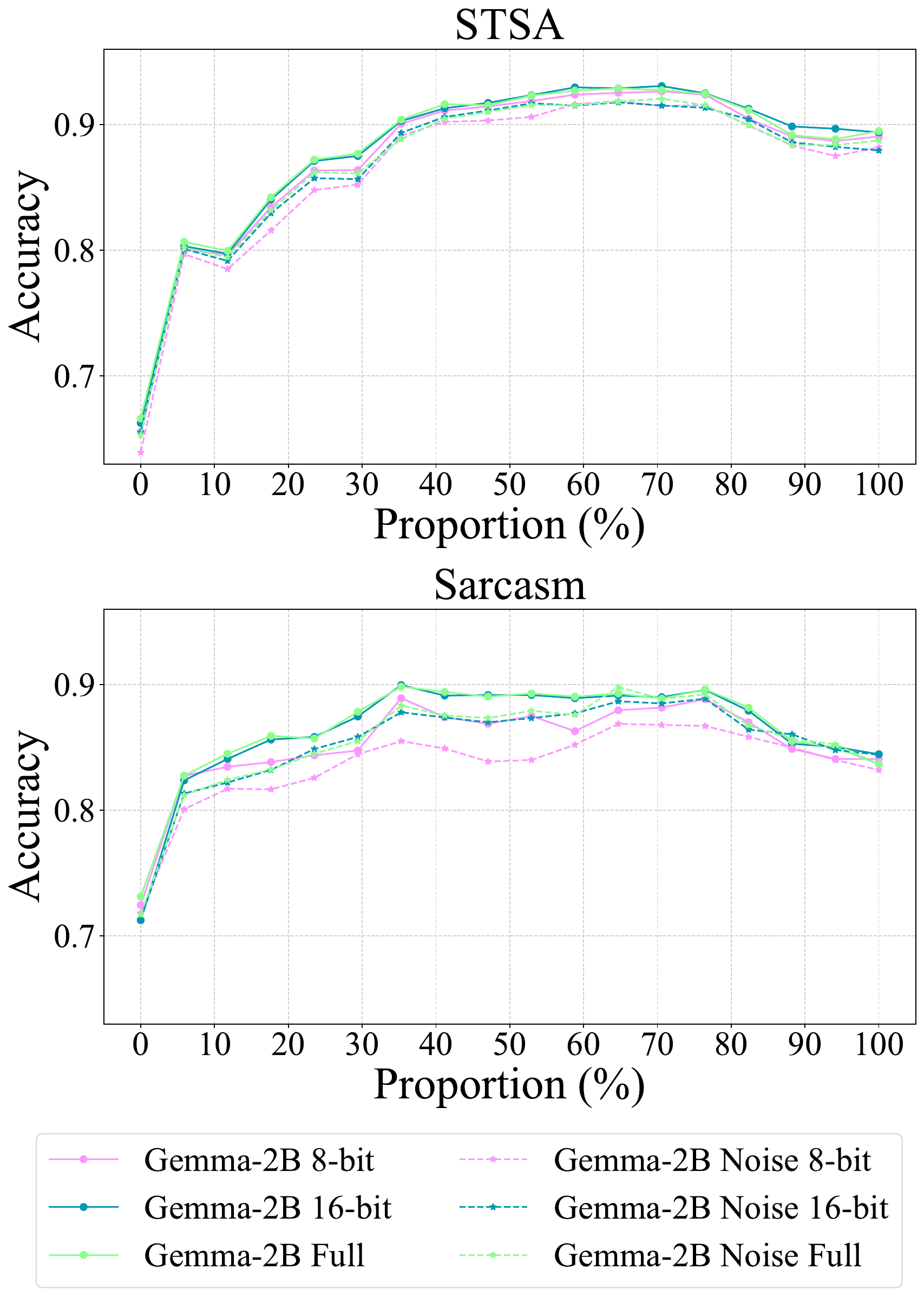}
    \caption{Noise and quantization evaluation on STSA and Sarcasm of Gemma-2B.}
    %\vspace{0.3cm}
    \label{fig:5342}
\end{center}
\end{figure}

\vspace{0.1cm}
\noindent \textbf{Results.} \autoref{fig:5342} illustrates the effect of adding noise and reducing bit representations to the Gemma-2B model. The addition of noise causes the learning curve to shift to the right, indicating a reduction in the converging speed. This suggests that the presence of noise in the input data can hinder the LLM's learning, slowing down its ability to converge to an optimal solution. We also see that there isn’t much difference between 32 and 16 bits, and the convergence rate slows when we reduce the model to 8 bits. Therefore, we may compress the LLMs to 16 bits for future designs.

\begin{tcolorbox}[colback=gray!10!white,colframe=gray!70!black,boxrule=0.3pt,title=Remark 4]
Noises or 8-bit-quantization can cause the accuracy to converge more slowly. Compressing the LLMs to 16 bits doesn't harm the understanding process too much.
The layer-wise representations of LLMs are susceptible to noise and high-ratio quantization. Therefore, it is crucial to proceed cautiously when conducting high-ratio quantization inference.
\end{tcolorbox}
\subsection{Examples of Each Dataset}
\label{sub:exampledata}

For the 9 bi-classification tasks, we provide two examples from each class for each task. The blue sentences are the data samples, while the black sentences are the prompts.

% Fill in the contents "prompt + question" for each dataset, the content that is actually fed by LLMs.
\vspace{0.06cm}
\noindent \textbf{\large Cities} 
\vspace{0.06cm}

\noindent \texttt{``True'' example:}

\noindent \texttt{Judge the statement is True or False.} \texttt{\color{blue} The city of Tokyo is in Japan.}

\vspace{0.06cm}

\noindent \texttt{``False'' example:}

\noindent \texttt{Judge the statement is True or False.} \texttt{\color{blue} The city of Lodz is in the Dominican Republic.}

\vspace{0.06cm}
\noindent \textbf{\large CommonClaim} 
\vspace{0.06cm}

\noindent \texttt{``True'' example:}

\noindent \texttt{Judge the statement is True or False.} \texttt{\color{blue} Salmon will often return to the same spawning ground where then were born.}

\vspace{0.06cm}

\noindent \texttt{``False'' example:}

\noindent \texttt{Judge the statement is True or False.} \texttt{\color{blue} A chicken has two right wings.}

\vspace{0.06cm}
\noindent \textbf{\large Counterfact} 
\vspace{0.06cm}

\noindent \texttt{``True'' example:}

\noindent \texttt{Judge the statement is True or False. } \texttt{\color{blue} The city of Tokyo is in Japan.}

\vspace{0.06cm}

\noindent \texttt{``False'' example:}

\noindent \texttt{Judge the statement is True or False.} \texttt{\color{blue} Kanata South Ward is in Wisconsin.}

\vspace{0.06cm}
\noindent  \textbf{\large HateEval}
\vspace{0.06cm}

\noindent \texttt{``Yes'' example:}

\noindent \texttt{\color{blue} Here it is not about Refugees or Illegal immigrants. It is about whether one has documents before 1971 or not. Now, it is difficult for slum people and beggars to show valid documents, except the name in voter list.} \texttt{According to the comment, tell whether they present hate speech or not.}  

\vspace{0.06cm}

\noindent \texttt{``No'' example:}

\noindent \texttt{\color{blue} Labor migrants transfer almost \$10 billion a year to Ukraine.} \texttt{According to the comment, tell whether they present hate speech or not.} 

\vspace{0.06cm}
\noindent \textbf{\large STSA}
\vspace{0.06cm}

\noindent \texttt{``Positive'' example:}

\noindent \texttt{\color{blue} The production values are of the highest and the performances attractive without being memorable.} \texttt{The sentence above is a movie review and reflects the writer's overall intention for this review. According to the sentence, judge whether the emotion is Positive or Negative.}
\vspace{0.06cm}

\noindent \texttt{``Negative'' example:}

\noindent \texttt{\color{blue} Less a story than an inexplicable nightmare, right down to the population's shrugging acceptance to each new horror.} \texttt{The sentence above is a movie review and reflects the writer's overall intention for this review. According to the sentence, judge whether the emotion is Positive or Negative.}

\vspace{0.06cm}
\noindent \textbf{\large IMDb}
\vspace{0.06cm}

\noindent \texttt{``Positive'' example:}

\noindent \texttt{\color{blue} This is the definitive movie version of Hamlet. Branagh cuts nothing, but there are no wasted moments.} \texttt{According to the movie review, judge whether it is Positive or Negative. }
\vspace{0.06cm}

\noindent \texttt{``Negative'' example:}

\noindent \texttt{\color{blue} This is without a doubt the worst movie I have ever seen. It is not funny. It is not interesting and should not have been made.} \texttt{According to the movie review, judge whether it is Positive or Negative. }

\vspace{0.06cm}
\noindent \textbf{\large Sarcasm}
\vspace{0.06cm}

\noindent \texttt{``Yes'' example:}

\noindent \texttt{Task: Detect sarcasm, help me identify whether this sentence is sarcastic. First, we need to understand what sarcasm is. Sarcasm is a form of verbal irony, where the intended meaning of the words is the opposite of the literal meaning. In other words, the speaker is saying one thing but meaning the opposite. }\texttt{\color{blue} Bashar al-Assad tries a tiny bit of sarin gas on self to see what it's like.}  \texttt{Think carefully according to the sentence. Is there any sarcasm in this sentence? Please answer Yes or No.}
\vspace{0.06cm}

\noindent \texttt{``No'' example:}

\noindent \texttt{Task: Detect sarcasm, help me identify whether this sentence is sarcastic. First, we need to understand what sarcasm is. Sarcasm is a form of verbal irony, where the intended meaning of the words is the opposite of the literal meaning. In other words, the speaker is saying one thing but meaning the opposite. } \texttt{\color{blue} This ceo will send your kids to school, if you work for his company.} \texttt{Think carefully according to the sentence. Is there any sarcasm in this sentence? Please answer Yes or No.}

\vspace{0.06cm}
\noindent \textbf{\large StrategyQA} 
\vspace{0.06cm}

\noindent Note: This dataset was created in 2021. Queen Elizabeth was alive then.

\noindent \texttt{``Yes'' example:}

\noindent \texttt{Judge the question is true or false? Q: Will Queen Elizabeth be buried in the Pantheon? Let us think step by step. The stem of the sentence is Queen Elizabeth, burial, pantheon. Inference: First, the Pantheon is a church, so it is possible that she could be buried there. Second, Queen Elizabeth II is still alive, so she has not been buried yet. Third, even if she were to be buried in the Pantheon, it is unlikely that we would know about it ahead of time, so it is hard to say for sure. pred\_ans: no.}  \texttt{\color{blue} Do hamsters provide food for any animals? } \texttt{Let us think step by step...}
\vspace{0.06cm}

\noindent \texttt{``No'' example:}

\noindent \texttt{Judge the question is true or false? Q: Will Queen Elizabeth be buried in the Pantheon? Let us think step by step. The stem of the sentence is Queen Elizabeth, burial, pantheon. Inference: First, the Pantheon is a church, so it is possible that she could be buried there. Second, Queen Elizabeth II is still alive, so she has not been buried yet. Third, even if she were to be buried in the Pantheon, it is unlikely that we would know about it ahead of time, so it is hard to say for sure. pred\_ans: no.}  \texttt{\color{blue}Could a llama birth twice during the War in Vietnam (1945-46)? } \texttt{Let us think step by step...}

\vspace{0.06cm}
\noindent \textbf{\large Coinflip}
\vspace{0.06cm}

\noindent \texttt{``Yes'' example:}

\noindent \texttt{\color{blue} A coin is heads up. Whitney flips the coin. Erika does not flip the coin. Tj does not flip the coin. Benito flips the coin. Is the coin still heads up? Note that "flip" here means "reverse".} \texttt{According to the flipping process above, determine if a coin remains heads up after it is either flipped or left unflipped by individuals. Therefore, the answer (Yes or No) is?}

\noindent \texttt{``No'' example:}

\noindent \texttt{\color{blue}A coin is heads up. Lucky does not flip the coin. Mireya flips the coin. Jj flips the coin. Kc flips the coin. Is the coin still heads up? Note that "flip" here means "reverse".} \texttt{According to the flipping process above, determine if a coin remains heads up after it is either flipped or left unflipped by individuals. Therefore, the answer (Yes or No) is?}

\clearpage

\begin{table}[tb]
\scriptsize
\renewcommand\arraystretch{1}

\centering
\resizebox{0.26\textwidth}{!}{
\begin{tabular}{l l c l}
\toprule
& Category & Dataset & \\
\midrule
 & \multirow{3}{*}{Fact} & Cities &  \\
 &  & Common &  \\
 &  & Counterfact &  \\
 \midrule
 & \multirow{4}{*}{Emotion} & HateEval &  \\
 &  & STSA &  \\
 &  & IMDb &  \\
 &  & Sarcasm &  \\
 \midrule
 & \multirow{2}{*}{Reasoning} & StrategyQA &  \\
 &  & Coinflip &  \\
\bottomrule
\end{tabular}}
\caption{The category that each dataset belongs to.}
\label{tab:datasetcate}

\end{table}

\vspace{-10pt}

\begin{table}[tb]
\scriptsize
\renewcommand\arraystretch{1}

\setlength{\tabcolsep}{1.5pt}
\centering
\begin{tabular}{@{}ccccc@{}}
\toprule
\textbf{Dataset}     & \textbf{LLaMA3-8B-Instruct} & \textbf{GPT-4o-mini} & \textbf{QWen2-7B-Instruct} & \textbf{Average} \\ \midrule
\textbf{Coinflip}    & 0.5080                       & 0.7620                & 0.5060                      & \cellcolor{cyan!31}0.5920           \\ 
\textbf{Common}      & 0.5606                      & 0.6905               & 0.6950                      & \cellcolor{cyan!35}0.6487           \\
\textbf{Sarcasm}     & 0.6575                      & 0.6770                & 0.6445                     & \cellcolor{cyan!36}0.6597           \\
\textbf{StrategyQA}  & 0.7035                      & 0.8803               & 0.5069                     & \cellcolor{cyan!46}0.6969           \\
\textbf{Counterfact} & 0.5277                      & 0.7990                & 0.8110                      & \cellcolor{cyan!50}0.7126           \\
\textbf{Hateeval}    & 0.7640                       & 0.7300                 & 0.7952                     & \cellcolor{cyan!56}0.7631           \\
\textbf{STSA}        & 0.9030                       & 0.9211               & 0.9108                     & \cellcolor{cyan!69}0.9116           \\
\textbf{Cities}      & 0.7687                      & \textbf{0.9973}      & \textbf{0.9953}            & \cellcolor{cyan!70}0.9204           \\
\textbf{IMDb}        & \textbf{0.9365}             & 0.9370                & 0.9405                     & \cellcolor{cyan!74}0.9380  \\
\bottomrule
\end{tabular}
\caption{Accuracy on nine datasets based on LLaMA3-8b-Instruct, GPT-4o-mini and QWen2-7B-Instruct.}
\label{tab:detailanchor}

\end{table}

\subsection{LLM structure}

Here, we give an introduction to the model structure (using LLaMA2-7B as an example).
\begin{verbatim}
LlamaForCausalLM(
    (model): LlamaModel(
        (embed_tokens):Embedding(32000,4096)
            (layers):ModuleList(
                (0-31):32 x lamaDecoderLayer(
                    (self_attn): LlamaSdpaAttention(
                        (g_proj):Linear(in_features=4096,out_features=4096,bias=False)
                        (k proj): Linear(in_features=4096,out_features=4096,bias=False)
                        (v proj):Linear(in_features=4096,out_features=4096, bias=False)
                        (o proj):Linear(in_features=4096, out_features=4096, bias=False)
                        (rotary emb):LlamaRotaryEmbedding()
                    )
                    (mlp):LlamaMLP(
                        (gate_proj): Linear(in_features=4096,out_features=11008,bias=False)
                        (up_proj): Linear(in_features=4096,out_features=11008,bias=False)
                        (down_proj): Linear(in_features=11008, out_features=4096, bias=False)
                        (act_fn): siLu()
                    )
                    (input_layernorm):LlamaRMSNorm()
                    (post_attention_layernorm):LlamaRMSNorm()
                )
            )
            (norm):LlamaRMSNorm()
        )
   (lm_head): Linear(in_features=4096,out_features=32000,bias=False)
)
\end{verbatim}
\end{appendices}

\end{document}

%%%%%%%%%%%%%%%%%%%%%%%%%%%%%%%%%%%%%%%%%%%%%%%%%%%%%%%%%%%%%%%%%%%%%%%%

%%% Use this command to specify your submission number.
%%% In doubleblind mode, it will be printed on the first page.

\paperid{123} 

%%% Use this command to specify the title of your paper.

\title{Exploring Concept Depth: How Large Language Models Acquire Knowledge at Different Layers?}

\author{Anonymous et al.}
\address{Anonymous Institution}
%%% Use this environment to include an abstract of your paper.
\begin{appendices}
\section{Appendix}

Here, we provide our supplementary materials.

\subsection{Metrics for Parts of the Layers}

\autoref{tab:stat} shows the experimental results for the accuracy, F1-score, and AUC metrics of parts of the first, 25\% depth, 50\% depth, 67\% depth, 83\% depth, and the last layer of each model over the nine datasets.

\begin{table*}[tb]
\renewcommand\arraystretch{1.6}
\centering
\caption{The experimental results of the nine datasets on nine LLMs. For each LLM, we select six layers (first, 25\% depth, 50\% depth, 67\% depth, 83\% depth, last) to record the accuracy, F1-score, and AUC.}
\resizebox{1.0\textwidth}{!}{%
\begin{tabular}{l|ccc|ccc|ccc|ccc|ccc|ccc|ccc|ccc|ccc}
    \toprule
Gemma-2b (18 Layers) & \multicolumn{3}{|c|}{StrategyQA } & \multicolumn{3}{|c|}{Coin} & \multicolumn{3}{|c|}{Cities}&\multicolumn{3}{|c|}{Common Claim}&\multicolumn{3}{|c|}{Counterfact}&\multicolumn{3}{|c|}{HateEval}&\multicolumn{3}{|c|}{STSA}&\multicolumn{3}{|c|}{IMDb}&\multicolumn{3}{|c}{Sarcasm}\\
\hline
Metrics &ACC & AUC  & F1 &ACC & AUC  & F1 &ACC & AUC  & F1 &ACC & AUC  & F1 &ACC & AUC  & F1 &ACC & AUC  & F1 &ACC & AUC  & F1 &ACC & AUC  & F1 &ACC & AUC  & F1   \\
\hline
1st-layer&0.556&0.601&0.588&0.635&0.667&0.626&0.446&0.411&0.422&0.582&0.61&0.584&0.502&0.509&0.518&0.74&0.822&0.742&0.666&0.729&0.68&0.722&0.788&0.725&0.731&0.808&0.714\\
25\%-layer&0.602&0.642&0.629&0.62&0.64&0.6&0.94&0.987&0.939&0.637&0.683&0.633&0.675&0.745&0.664&0.793&0.89&0.794&0.872&0.942&0.874&0.884&0.953&0.885&0.857&0.935&0.852\\
50\%-layer&0.639&0.7&0.665&0.65&0.705&0.632&0.983&0.999&0.983&0.648&0.699&0.642&0.729&0.826&0.716&0.802&0.891&0.8&0.915&0.972&0.917&0.941&0.982&0.941&0.89&0.955&0.887\\
67\%-layer&0.683&0.751&0.708&0.695&0.783&0.69&0.992&1.0&0.992&0.671&0.727&0.666&0.766&0.852&0.758&0.809&0.891&0.806&0.929&0.976&0.93&0.936&0.984&0.936&0.893&0.96&0.889\\
83\%-layer&0.62&0.668&0.637&0.585&0.625&0.579&0.985&0.999&0.985&0.652&0.703&0.645&0.704&0.782&0.696&0.807&0.89&0.803&0.911&0.97&0.914&0.926&0.977&0.926&0.882&0.944&0.878\\
last-layer&0.592&0.602&0.626&0.525&0.554&0.532&0.988&0.999&0.988&0.647&0.697&0.644&0.685&0.752&0.679&0.803&0.892&0.801&0.895&0.957&0.898&0.941&0.981&0.941&0.837&0.92&0.831\\

\hline
Gemma-7b (28 Layers) & \multicolumn{3}{|c|}{StrategyQA } & \multicolumn{3}{|c|}{Coin} & \multicolumn{3}{|c|}{Cities}&\multicolumn{3}{|c|}{Common Claim}&\multicolumn{3}{|c|}{Counterfact}&\multicolumn{3}{|c|}{HateEval}&\multicolumn{3}{|c|}{STSA}&\multicolumn{3}{|c|}{IMDb}&\multicolumn{3}{|c}{Sarcasm}\\
\hline
Metrics &ACC & AUC  & F1 &ACC & AUC  & F1 &ACC & AUC  & F1 &ACC & AUC  & F1 &ACC & AUC  & F1 &ACC & AUC  & F1 &ACC & AUC  & F1 &ACC & AUC  & F1 &ACC & AUC  & F1  \\
\hline

1st-layer&0.519&0.565&0.569&0.69&0.712&0.656&0.59&0.444&0.439&0.737&0.609&0.053&0.496&0.506&0.514&0.737&0.817&0.741&0.668&0.737&0.679&0.73&0.811&0.735&0.723&0.793&0.713\\
25\%-layer&0.617&0.653&0.648&0.605&0.654&0.591&0.95&0.987&0.951&0.719&0.687&0.422&0.696&0.773&0.686&0.795&0.875&0.794&0.89&0.958&0.893&0.9&0.958&0.899&0.885&0.953&0.882\\
50\%-layer&0.658&0.716&0.678&0.765&0.844&0.761&1.0&0.999&0.998&0.75&0.747&0.498&0.835&0.912&0.829&0.784&0.871&0.782&0.929&0.979&0.93&0.932&0.979&0.932&0.907&0.968&0.904\\
67\%-layer&0.733&0.809&0.744&0.88&0.922&0.875&1.0&0.998&0.995&0.756&0.75&0.515&0.814&0.9&0.809&0.784&0.868&0.782&0.936&0.981&0.937&0.922&0.979&0.923&0.913&0.972&0.91\\
83\%-layer&0.675&0.746&0.695&0.785&0.818&0.786&0.98&0.997&0.982&0.743&0.74&0.492&0.776&0.867&0.768&0.812&0.897&0.809&0.935&0.979&0.937&0.925&0.977&0.926&0.901&0.963&0.898\\
last-layer&0.604&0.666&0.625&0.57&0.602&0.578&0.95&0.996&0.972&0.759&0.748&0.481&0.729&0.808&0.721&0.82&0.901&0.817&0.92&0.975&0.922&0.938&0.984&0.938&0.862&0.932&0.86\\

\hline
LlaMA-7b (32 Layers) & \multicolumn{3}{|c|}{StrategyQA } & \multicolumn{3}{|c|}{Coin} & \multicolumn{3}{|c|}{Cities}&\multicolumn{3}{|c|}{Common Claim}&\multicolumn{3}{|c|}{Counterfact}&\multicolumn{3}{|c|}{HateEval}&\multicolumn{3}{|c|}{STSA}&\multicolumn{3}{|c|}{IMDb}&\multicolumn{3}{|c}{Sarcasm}\\
\hline
Metrics &ACC & AUC  & F1 &ACC & AUC  & F1 &ACC & AUC  & F1 &ACC & AUC  & F1 &ACC & AUC  & F1 &ACC & AUC  & F1 &ACC & AUC  & F1 &ACC & AUC  & F1 &ACC & AUC  & F1  \\
\hline

1st-layer&0.584&0.608&0.641&0.545&0.525&0.674&0.487&0.472&0.472&0.74&0.617&0.031&0.493&0.491&0.522&0.725&0.814&0.732&0.678&0.744&0.703&0.73&0.799&0.736&0.705&0.773&0.697\\
25\%-layer&0.657&0.712&0.688&0.615&0.612&0.621&0.93&0.978&0.929&0.736&0.699&0.441&0.684&0.754&0.681&0.806&0.887&0.792&0.513&0.913&0.676&0.914&0.972&0.914&0.89&0.961&0.886\\
50\%-layer&0.74&0.827&0.754&0.915&0.977&0.907&0.997&1.0&0.997&0.753&0.738&0.477&0.797&0.894&0.793&0.805&0.883&0.791&0.847&0.954&0.866&0.941&0.984&0.941&0.922&0.976&0.919\\
67\%-layer&0.729&0.805&0.744&0.9&0.966&0.89&0.995&1.0&0.995&0.744&0.729&0.466&0.775&0.872&0.77&0.795&0.88&0.779&0.896&0.957&0.902&0.939&0.985&0.939&0.92&0.973&0.918\\
83\%-layer&0.699&0.774&0.72&0.88&0.961&0.871&0.993&1.0&0.993&0.734&0.719&0.464&0.744&0.832&0.735&0.798&0.887&0.785&0.913&0.967&0.915&0.939&0.984&0.939&0.908&0.966&0.905\\
last-layer&0.67&0.744&0.69&0.815&0.9&0.8&0.993&1.0&0.993&0.743&0.731&0.464&0.739&0.818&0.731&0.831&0.914&0.83&0.935&0.984&0.937&0.944&0.987&0.944&0.895&0.955&0.892\\

\hline
LlaMA-13B (40 Layers) & \multicolumn{3}{|c|}{StrategyQA } & \multicolumn{3}{|c|}{Coin} & \multicolumn{3}{|c|}{Cities}&\multicolumn{3}{|c|}{Common Claim}&\multicolumn{3}{|c|}{Counterfact}&\multicolumn{3}{|c|}{HateEval}&\multicolumn{3}{|c|}{STSA}&\multicolumn{3}{|c|}{IMDb}&\multicolumn{3}{|c}{Sarcasm}\\
\hline
Metrics &ACC & AUC  & F1 &ACC & AUC  & F1 &ACC & AUC  & F1 &ACC & AUC  & F1 &ACC & AUC  & F1 &ACC & AUC  & F1 &ACC & AUC  & F1 &ACC & AUC  & F1 &ACC & AUC&F1   \\
\hline
1st-layer&0.567&0.601&0.628&0.475&0.509&0.575&0.481&0.463&0.47&0.741&0.62&0.034&0.486&0.485&0.526&0.719&0.807&0.727&0.697&0.769&0.709&0.732&0.795&0.736&0.692&0.756&0.688\\
25\%-layer&0.676&0.732&0.701&0.53&0.59&0.515&0.985&0.999&0.985&0.733&0.716&0.457&0.763&0.859&0.758&0.832&0.914&0.829&0.93&0.98&0.931&0.942&0.983&0.943&0.915&0.972&0.913\\
50\%-layer&0.763&0.844&0.771&0.825&0.886&0.819&0.993&1.0&0.993&0.758&0.751&0.515&0.812&0.897&0.809&0.839&0.92&0.836&0.939&0.984&0.94&0.945&0.984&0.945&0.936&0.983&0.934\\
67\%-layer&0.716&0.806&0.729&0.795&0.882&0.794&0.993&1.0&0.993&0.751&0.745&0.499&0.776&0.866&0.772&0.838&0.919&0.834&0.938&0.984&0.939&0.94&0.987&0.94&0.924&0.978&0.921\\
83\%-layer&0.71&0.795&0.719&0.7&0.797&0.703&0.99&1.0&0.99&0.741&0.731&0.487&0.768&0.856&0.762&0.832&0.912&0.829&0.937&0.983&0.938&0.941&0.985&0.942&0.922&0.974&0.919\\
last-layer&0.693&0.772&0.704&0.645&0.715&0.664&0.99&1.0&0.99&0.75&0.743&0.499&0.76&0.841&0.752&0.835&0.913&0.833&0.935&0.984&0.937&0.946&0.988&0.946&0.91&0.969&0.908\\

\hline
QWen-0.5B (24 Layers) & \multicolumn{3}{|c|}{StrategyQA } & \multicolumn{3}{|c|}{Coin} & \multicolumn{3}{|c|}{Cities}&\multicolumn{3}{|c|}{Common Claim}&\multicolumn{3}{|c|}{Counterfact}&\multicolumn{3}{|c|}{HateEval}&\multicolumn{3}{|c|}{STSA}&\multicolumn{3}{|c|}{IMDb}&\multicolumn{3}{|c}{Sarcasm}\\
\hline
Metrics &ACC & AUC  & F1 &ACC & AUC  & F1 &ACC & AUC  & F1 &ACC & AUC  & F1 &ACC & AUC  & F1 &ACC & AUC  & F1 &ACC & AUC  & F1 &ACC & AUC  & F1 &ACC & AUC&F1   \\
\hline

1st-layer&0.557&0.578&0.607&0.535&0.649&0.657&0.482&0.46&0.464&0.735&0.622&0.11&0.499&0.503&0.527&0.759&0.851&0.764&0.73&0.801&0.733&0.764&0.837&0.762&0.729&0.799&0.72\\
25\%-layer&0.583&0.63&0.62&0.545&0.582&0.508&0.731&0.797&0.722&0.732&0.651&0.257&0.52&0.524&0.523&0.785&0.864&0.783&0.751&0.829&0.756&0.804&0.887&0.806&0.811&0.895&0.799\\
50\%-layer&0.619&0.686&0.649&0.62&0.652&0.596&0.935&0.979&0.935&0.744&0.695&0.379&0.68&0.754&0.676&0.793&0.88&0.792&0.846&0.921&0.848&0.884&0.949&0.883&0.838&0.92&0.831\\
67\%-layer&0.644&0.688&0.673&0.585&0.617&0.561&0.933&0.982&0.934&0.742&0.705&0.375&0.668&0.74&0.665&0.789&0.874&0.786&0.868&0.946&0.87&0.894&0.956&0.893&0.827&0.911&0.821\\
83\%-layer&0.583&0.61&0.612&0.62&0.668&0.612&0.923&0.971&0.924&0.746&0.706&0.364&0.604&0.657&0.6&0.791&0.867&0.791&0.85&0.927&0.854&0.866&0.941&0.865&0.824&0.902&0.819\\
last-layer&0.55&0.567&0.584&0.55&0.613&0.541&0.912&0.971&0.912&0.742&0.703&0.357&0.579&0.616&0.579&0.784&0.866&0.781&0.844&0.922&0.848&0.879&0.951&0.88&0.825&0.9&0.82\\

\hline
QWen-1.8B (24 Layers) & \multicolumn{3}{|c|}{StrategyQA } & \multicolumn{3}{|c|}{Coin} & \multicolumn{3}{|c|}{Cities}&\multicolumn{3}{|c|}{Common Claim}&\multicolumn{3}{|c|}{Counterfact}&\multicolumn{3}{|c|}{HateEval}&\multicolumn{3}{|c|}{STSA}&\multicolumn{3}{|c|}{IMDb}&\multicolumn{3}{|c}{Sarcasm}\\
\hline
Metrics &ACC & AUC  & F1 &ACC & AUC  & F1 &ACC & AUC  & F1 &ACC & AUC  & F1 &ACC & AUC  & F1 &ACC & AUC  & F1 &ACC & AUC  & F1 &ACC & AUC  & F1 &ACC & AUC&F1   \\
\hline

1st-layer&0.57&0.6&0.63&0.49&0.634&0.648&0.482&0.458&0.464&0.739&0.619&0.071&0.516&0.514&0.539&0.724&0.819&0.732&0.693&0.762&0.703&0.718&0.784&0.72&0.721&0.796&0.713\\
25\%-layer&0.607&0.638&0.643&0.58&0.59&0.584&0.583&0.626&0.582&0.736&0.658&0.317&0.521&0.541&0.525&0.809&0.882&0.807&0.775&0.844&0.781&0.81&0.899&0.81&0.833&0.909&0.825\\
50\%-layer&0.658&0.726&0.676&0.595&0.655&0.58&0.975&0.997&0.975&0.741&0.708&0.419&0.688&0.767&0.683&0.808&0.89&0.807&0.895&0.961&0.897&0.914&0.974&0.915&0.87&0.947&0.864\\
67\%-layer&0.664&0.733&0.685&0.695&0.759&0.655&0.977&0.996&0.977&0.741&0.717&0.423&0.695&0.776&0.689&0.809&0.886&0.804&0.893&0.963&0.895&0.904&0.968&0.905&0.865&0.943&0.859\\
83\%-layer&0.631&0.666&0.649&0.7&0.747&0.674&0.972&0.995&0.972&0.735&0.706&0.419&0.657&0.734&0.651&0.791&0.877&0.788&0.89&0.956&0.893&0.891&0.957&0.893&0.835&0.922&0.829\\
last-layer&0.595&0.642&0.604&0.615&0.667&0.605&0.973&0.996&0.973&0.74&0.704&0.404&0.638&0.713&0.631&0.795&0.877&0.794&0.879&0.949&0.882&0.906&0.962&0.907&0.812&0.896&0.808\\

\hline
QWen-7B (40 Layers)& \multicolumn{3}{|c|}{StrategyQA } & \multicolumn{3}{|c|}{Coin} & \multicolumn{3}{|c|}{Cities}&\multicolumn{3}{|c|}{Common Claim}&\multicolumn{3}{|c|}{Counterfact}&\multicolumn{3}{|c|}{HateEval}&\multicolumn{3}{|c|}{STSA}&\multicolumn{3}{|c|}{IMDb}&\multicolumn{3}{|c}{Sarcasm}\\
\hline
Metrics &ACC & AUC  & F1 &ACC & AUC  & F1 &ACC & AUC  & F1 &ACC & AUC  & F1 &ACC & AUC  & F1 &ACC & AUC  & F1 &ACC & AUC  & F1 &ACC & AUC  & F1 &ACC & AUC&F1   \\
\hline

1st-layer&0.54&0.559&0.602&0.545&0.597&0.588&0.437&0.403&0.418&0.735&0.635&0.169&0.475&0.466&0.498&0.785&0.867&0.786&0.753&0.826&0.757&0.782&0.858&0.785&0.771&0.856&0.765\\
25\%-layer&0.59&0.625&0.631&0.62&0.635&0.631&0.806&0.89&0.803&0.721&0.657&0.348&0.556&0.57&0.553&0.798&0.879&0.795&0.782&0.869&0.787&0.825&0.916&0.827&0.852&0.924&0.848\\
50\%-layer&0.705&0.782&0.724&0.66&0.719&0.667&0.985&0.998&0.985&0.744&0.733&0.452&0.731&0.825&0.722&0.802&0.885&0.795&0.912&0.969&0.914&0.929&0.975&0.929&0.882&0.956&0.878\\
67\%-layer&0.702&0.773&0.722&0.635&0.719&0.64&0.983&0.997&0.983&0.746&0.728&0.465&0.721&0.801&0.716&0.793&0.88&0.789&0.904&0.965&0.907&0.915&0.968&0.913&0.89&0.955&0.888\\
83\%-layer&0.678&0.751&0.703&0.685&0.762&0.693&0.978&0.996&0.978&0.747&0.724&0.45&0.68&0.747&0.667&0.795&0.881&0.79&0.899&0.961&0.901&0.916&0.965&0.915&0.872&0.941&0.868\\
last-layer&0.676&0.718&0.693&0.585&0.623&0.587&0.982&0.995&0.981&0.75&0.727&0.453&0.657&0.718&0.643&0.782&0.869&0.778&0.902&0.964&0.905&0.92&0.973&0.919&0.848&0.921&0.845\\

\hline
QWen-14B (32 Layers) & \multicolumn{3}{|c|}{StrategyQA } & \multicolumn{3}{|c|}{Coin} & \multicolumn{3}{|c|}{Cities}&\multicolumn{3}{|c|}{Common Claim}&\multicolumn{3}{|c|}{Counterfact}&\multicolumn{3}{|c|}{HateEval}&\multicolumn{3}{|c|}{STSA}&\multicolumn{3}{|c|}{IMDb}&\multicolumn{3}{|c}{Sarcasm}\\
\hline
Metrics &ACC & AUC  & F1 &ACC & AUC  & F1 &ACC & AUC  & F1 &ACC & AUC  & F1 &ACC & AUC  & F1 &ACC & AUC  & F1 &ACC & AUC  & F1 &ACC & AUC  & F1 &ACC & AUC&F1   \\
\hline

1st-layer&0.582&0.608&0.624&0.555&0.599&0.594&0.436&0.399&0.429&0.738&0.627&0.205&0.487&0.481&0.491&0.77&0.864&0.772&0.749&0.827&0.754&0.768&0.864&0.768&0.785&0.866&0.773\\
25\%-layer&0.605&0.645&0.644&0.555&0.609&0.566&0.858&0.93&0.855&0.713&0.667&0.362&0.558&0.601&0.555&0.8&0.879&0.796&0.818&0.898&0.821&0.861&0.936&0.864&0.876&0.946&0.871\\
50\%-layer&0.695&0.775&0.709&0.69&0.731&0.69&0.99&0.999&0.99&0.744&0.742&0.483&0.771&0.861&0.762&0.812&0.897&0.807&0.918&0.972&0.919&0.946&0.983&0.946&0.901&0.966&0.897\\
67\%-layer&0.734&0.82&0.749&0.685&0.753&0.693&0.992&0.997&0.992&0.762&0.754&0.5&0.769&0.848&0.765&0.802&0.889&0.798&0.914&0.974&0.916&0.93&0.978&0.93&0.898&0.963&0.895\\
83\%-layer&0.737&0.814&0.751&0.68&0.74&0.677&0.992&0.996&0.992&0.751&0.732&0.485&0.726&0.808&0.719&0.806&0.886&0.805&0.91&0.972&0.912&0.925&0.974&0.925&0.892&0.957&0.887\\
last-layer&0.714&0.785&0.729&0.65&0.681&0.646&0.982&0.996&0.981&0.753&0.74&0.483&0.707&0.779&0.701&0.802&0.881&0.798&0.908&0.97&0.91&0.932&0.979&0.933&0.871&0.943&0.867\\

\hline
QWen-14B (40 Layers) & \multicolumn{3}{|c|}{StrategyQA } & \multicolumn{3}{|c|}{Coin} & \multicolumn{3}{|c|}{Cities}&\multicolumn{3}{|c|}{Common Claim}&\multicolumn{3}{|c|}{Counterfact}&\multicolumn{3}{|c|}{HateEval}&\multicolumn{3}{|c|}{STSA}&\multicolumn{3}{|c|}{IMDb}&\multicolumn{3}{|c}{Sarcasm}\\
\hline
Metrics &ACC & AUC  & F1 &ACC & AUC  & F1 &ACC & AUC  & F1 &ACC & AUC  & F1 &ACC & AUC  & F1 &ACC & AUC  & F1 &ACC & AUC  & F1 &ACC & AUC  & F1 &ACC & AUC&F1   \\
\hline
1st-layer&0.569&0.59&0.626&0.605&0.639&0.646&0.461&0.422&0.448&0.733&0.62&0.162&0.481&0.486&0.504&0.766&0.86&0.769&0.75&0.824&0.753&0.778&0.856&0.776&0.772&0.859&0.762\\
25\%-layer&0.603&0.637&0.631&0.68&0.704&0.66&0.866&0.945&0.865&0.72&0.669&0.393&0.621&0.67&0.615&0.8&0.89&0.797&0.842&0.919&0.845&0.862&0.939&0.861&0.874&0.946&0.869\\
50\%-layer&0.735&0.835&0.747&0.71&0.754&0.681&0.995&1.0&0.995&0.76&0.748&0.515&0.791&0.875&0.787&0.808&0.891&0.807&0.932&0.981&0.934&0.935&0.984&0.935&0.919&0.974&0.916\\
67\%-layer&0.798&0.883&0.808&0.7&0.791&0.674&0.995&1.0&0.995&0.761&0.76&0.502&0.775&0.859&0.769&0.816&0.894&0.814&0.933&0.98&0.934&0.938&0.978&0.937&0.914&0.968&0.911\\
83\%-layer&0.796&0.872&0.807&0.715&0.768&0.705&0.992&1.0&0.992&0.753&0.735&0.493&0.753&0.843&0.748&0.816&0.896&0.811&0.927&0.978&0.927&0.936&0.978&0.936&0.912&0.964&0.91\\
last-layer&0.783&0.863&0.792&0.605&0.67&0.586&0.99&1.0&0.99&0.757&0.749&0.481&0.725&0.814&0.723&0.815&0.897&0.812&0.922&0.978&0.923&0.931&0.978&0.93&0.89&0.952&0.887\\

    \bottomrule

\end{tabular}%
}
\label{tab:stat}

\end{table*}

\subsection{Examples of Each Dataset}
\label{sub:exampledata}

For the 9 bi-classification tasks, we provide two examples from each class for each task. The blue sentences are the data samples, while the black sentences are the prompts.

% Fill in the contents "prompt + question" for each dataset, the content that is actually fed by LLMs.
\vspace{0.06cm}
\noindent \textbf{\large Cities} 
\vspace{0.06cm}

\noindent \texttt{"True" example:}

\noindent \texttt{Judge the statement is True or False.} \texttt{\color{blue} The city of Tokyo is in Japan.}

\vspace{0.06cm}

\noindent \texttt{"False" example:}

\noindent \texttt{Judge the statement is True or False.} \texttt{\color{blue} The city of Lodz is in the Dominican Republic.}

\vspace{0.06cm}
\noindent \textbf{\large CommonClaim} 
\vspace{0.06cm}

\noindent \texttt{"True" example:}

\noindent \texttt{Judge the statement is True or False.} \texttt{\color{blue} Salmon will often return to the same spawning ground where then were born.}

\vspace{0.06cm}

\noindent \texttt{"False" example:}

\noindent \texttt{Judge the statement is True or False.} \texttt{\color{blue} A chicken has two right wings.}

\vspace{0.06cm}
\noindent \textbf{\large Counterfact} 
\vspace{0.06cm}

\noindent \texttt{"True" example:}

\noindent \texttt{Judge the statement is True or False. } \texttt{\color{blue} The city of Tokyo is in Japan.}

\vspace{0.06cm}

\noindent \texttt{"False" example:}

\noindent \texttt{Judge the statement is True or False.} \texttt{\color{blue} Kanata South Ward is in Wisconsin.}

\vspace{0.06cm}
\noindent  \textbf{\large HateEval}
\vspace{0.06cm}

\noindent \texttt{"Yes" example:}

\noindent \texttt{\color{blue} Here it is not about Refugees or Illegal immigrants. It is about whether one has documents before 1971 or not. Now, it is difficult for slum people and beggars to show valid documents, except the name in voter list.} \texttt{According to the comment, tell whether they present hate speech or not.}  

\vspace{0.06cm}

\noindent \texttt{"No" example:}

\noindent \texttt{\color{blue} Labor migrants transfer almost \$10 billion a year to Ukraine.} \texttt{According to the comment, tell whether they present hate speech or not.} 

\vspace{0.06cm}
\noindent \textbf{\large STSA}
\vspace{0.06cm}

\noindent \texttt{"Positive" example:}

\noindent \texttt{\color{blue} The production values are of the highest and the performances attractive without being memorable.} \texttt{The sentence above is a movie review and reflects the writer's overall intention for this review. According to the sentence, judge whether the emotion is Positive or Negative.}
\vspace{0.06cm}

\noindent \texttt{"Negative" example:}

\noindent \texttt{\color{blue} Less a story than an inexplicable nightmare, right down to the population's shrugging acceptance to each new horror.} \texttt{The sentence above is a movie review and reflects the writer's overall intention for this review. According to the sentence, judge whether the emotion is Positive or Negative.}

\vspace{0.06cm}
\noindent \textbf{\large IMDb}
\vspace{0.06cm}

\noindent \texttt{"Positive" example:}

\noindent \texttt{\color{blue} This is the definitive movie version of Hamlet. Branagh cuts nothing, but there are no wasted moments.} \texttt{According to the movie review, judge whether it is Positive or Negative. }
\vspace{0.06cm}

\noindent \texttt{"Negative" example:}

\noindent \texttt{\color{blue} This is without a doubt the worst movie I have ever seen. It is not funny. It is not interesting and should not have been made.} \texttt{According to the movie review, judge whether it is Positive or Negative. }

\vspace{0.06cm}
\noindent \textbf{\large Sarcasm}
\vspace{0.06cm}

\noindent \texttt{"Yes" example:}

\noindent \texttt{Task: Detect sarcasm, help me identify whether this sentence is sarcastic. First, we need to understand what sarcasm is. Sarcasm is a form of verbal irony, where the intended meaning of the words is the opposite of the literal meaning. In other words, the speaker is saying one thing but meaning the opposite. }\texttt{\color{blue} Bashar al-Assad tries a tiny bit of sarin gas on self to see what it's like.}  \texttt{Think carefully according to the sentence. Is there any sarcasm in this sentence? Please answer Yes or No.}
\vspace{0.06cm}

\noindent \texttt{"No" example:}

\noindent \texttt{Task: Detect sarcasm, help me identify whether this sentence is sarcastic. First, we need to understand what sarcasm is. Sarcasm is a form of verbal irony, where the intended meaning of the words is the opposite of the literal meaning. In other words, the speaker is saying one thing but meaning the opposite. } \texttt{\color{blue} This ceo will send your kids to school, if you work for his company.} \texttt{Think carefully according to the sentence. Is there any sarcasm in this sentence? Please answer Yes or No.}

\vspace{0.06cm}
\noindent \textbf{\large StrategyQA} 
\vspace{0.06cm}

\noindent Note: This dataset was created in 2021. Queen Elizabeth was alive then.

\noindent \texttt{"Yes" example:}

\noindent \texttt{Judge the question is true or false? Q: Will Queen Elizabeth be buried in the Pantheon? Let us think step by step. The stem of the sentence is Queen Elizabeth, burial, pantheon. Inference: First, the Pantheon is a church, so it is possible that she could be buried there. Second, Queen Elizabeth II is still alive, so she has not been buried yet. Third, even if she were to be buried in the Pantheon, it is unlikely that we would know about it ahead of time, so it is hard to say for sure. pred\_ans: no.}  \texttt{\color{blue} Do hamsters provide food for any animals? } \texttt{Let us think step by step...}
\vspace{0.06cm}

\noindent \texttt{"No" example:}

\noindent \texttt{Judge the question is true or false? Q: Will Queen Elizabeth be buried in the Pantheon? Let us think step by step. The stem of the sentence is Queen Elizabeth, burial, pantheon. Inference: First, the Pantheon is a church, so it is possible that she could be buried there. Second, Queen Elizabeth II is still alive, so she has not been buried yet. Third, even if she were to be buried in the Pantheon, it is unlikely that we would know about it ahead of time, so it is hard to say for sure. pred\_ans: no.}  \texttt{\color{blue}Could a llama birth twice during the War in Vietnam (1945-46)? } \texttt{Let us think step by step...}

\vspace{0.06cm}
\noindent \textbf{\large Coinflip}
\vspace{0.06cm}

\noindent \texttt{"Yes" example:}

\noindent \texttt{\color{blue} A coin is heads up. Whitney flips the coin. Erika does not flip the coin. Tj does not flip the coin. Benito flips the coin. Is the coin still heads up? Note that "flip" here means "reverse".} \texttt{According to the flipping process above, determine if a coin remains heads up after it is either flipped or left unflipped by individuals. Therefore, the answer (Yes or No) is?}

\noindent \texttt{"No" example:}

\noindent \texttt{\color{blue}A coin is heads up. Lucky does not flip the coin. Mireya flips the coin. Jj flips the coin. Kc flips the coin. Is the coin still heads up? Note that "flip" here means "reverse".} \texttt{According to the flipping process above, determine if a coin remains heads up after it is either flipped or left unflipped by individuals. Therefore, the answer (Yes or No) is?}

\subsection{LLM structure}

Here, we give an introduction to the model structure (using LLaMA2-7B as an example).
\begin{verbatim}
LlamaForCausalLM(
    (model): LlamaModel(
        (embed_tokens):Embedding(32000,4096)
            (layers):ModuleList(
                (0-31):32 x lamaDecoderLayer(
                    (self_attn): LlamaSdpaAttention(
                        (g_proj):Linear(in_features=4096,out_features=4096,bias=False)
                        (k proj): Linear(in_features=4096,out_features=4096,bias=False)
                        (v proj):Linear(in_features=4096,out_features=4096, bias=False)
                        (o proj):Linear(in_features=4096, out_features=4096, bias=False)
                        (rotary emb):LlamaRotaryEmbedding()
                    )
                    (mlp):LlamaMLP(
                        (gate_proj): Linear(in_features=4096,out_features=11008,bias=False)
                        (up_proj): Linear(in_features=4096,out_features=11008,bias=False)
                        (down_proj): Linear(in_features=11008, out_features=4096, bias=False)
                        (act_fn): siLu()
                    )
                    (input_layernorm):LlamaRMSNorm()
                    (post_attention_layernorm):LlamaRMSNorm()
                )
            )
            (norm):LlamaRMSNorm()
        )
   (lm_head): Linear(in_features=4096,out_features=32000,bias=False)
)
\end{verbatim}
% \subsection{$d_{model}$ of each LLM}
% %\vspace{0.3cm}
% \renewcommand\arraystretch{1.5}
% \centering
% \begin{tabular}{| l | l | l | l | l | l |}
% \hline
% Model & $d_{model}$ & Model & $d_{model}$ & Model & $d_{model}$\\
% \hline
% Gemma-2B & 18 & Gemma-7B & 28 & LLaMA-7B & 4096 \\
% \hline
% LLaMA-13B & 4096 & QWen-0.5B & 24 & QWen-1.8B & 24 \\
% \hline
% QWen-4B & 40 & QWen-7B & 32 & QWen-14B & 40 \\
% \hline

% \end{tabular}
    
% \label{tab:numlayer}

%%%%%%%%%%%%%%%%%%%%%%%%%%%%%%%%%%%%%%%%%%%%%%%%%%%%%%%%%%%%%%%%%%%%%%%%

%%%%%%%%%%%%%%%%%%%%%%%%%%%%%%%%%%%%%%%%%%%%%%%%%%%%%%%%%%%%%%%%%%%%%%%%

%%% Use this environment to include acknowledgements (optional).
%%% This will be omitted in doubleblind mode.

% \begin{ack}
% By using the ack} environment to insert your (optional) 
% acknowledgements, you can ensure that the text is suppressed whenever 
% you use the doubleblind} option. In the final version, 
% acknowledgements may be included on the extra page intended for references.
% \end{ack}

%%%%%%%%%%%%%%%%%%%%%%%%%%%%%%%%%%%%%%%%%%%%%%%%%%%%%%%%%%%%%%%%%%%%%%%%
\end{appendices}

%%% Use this command to include your bibliography file.